\documentclass[preprint,12pt]{elsarticle}

\usepackage{amssymb}
\usepackage{amsmath}
\usepackage[hidelinks]{hyperref}
\usepackage{caption,tabularx,booktabs, array, makecell, subcaption, amsmath}
\usepackage{appendix}
\usepackage{orcidlink}
\usepackage{float}
\newcommand{\PreserveBackslash}[1]{\let\temp=\\#1\let\\=\temp}
\newcolumntype{C}[1]{>{\PreserveBackslash\centering}p{#1}}
\newcolumntype{R}[1]{>{\PreserveBackslash\raggedleft}p{#1}}
\newcolumntype{L}[1]{>{\PreserveBackslash\raggedright}p{#1}}

\journal{}

\begin{document}

\begin{frontmatter}



\title{Evaluating Hierarchical Clinical Document Classification Using Reasoning-Based LLMs\tnoteref{Evaluating Hierarchical Clinical Document Classification Using Reasoning-Based LLMs}}
% \tnotetext[Human vs Large Language Models for Classifying Clinical Documents]{}
 
\author[label1]{Akram Mustafa\orcidlink{0000-0003-4090-2597}}
\ead{akram.mohdmustafa@my.jcu.edu.au}

\author[label2]{Usman Naseem\orcidlink{0000-0003-0191-7171}}
\ead{usman.naseem@mq.edu.au}

\author[label1]{Mostafa Rahimi Azghadi\orcidlink{0000-0001-7975-3985}}
\ead{mostafa.rahimiazghadi@jcu.edu.au}
\cortext[cor1]{mostafa.rahimiazghadi@jcu.edu.au}

 
 \affiliation[label1]{organization={College of Science and Engineering, James Cook University},
             city={Townsville},
             postcode={4811},
             state={QLD},
             country={Australia}}


 \affiliation[label2]{organization={School of Computing, Macquarie University},
             city={Sydney},
             postcode={2113},
             state={NSW},
             country={Australia}}


%% Abstract
\begin{abstract}
\noindent\textit{Background}: Clinical coding, particularly the classification of hierarchical ICD-10 codes from unstructured discharge summaries, is essential for healthcare operations, but remains a labor-intensive and error-prone task. Automated approaches using Large Language Models (LLMs) offer the potential to augment or replace human coders, yet their reliability and reasoning capabilities, which is needed to ensure accurate, explainable code assignments, are not well understood.\\
\noindent\textit{Objective}: This study aims to benchmark a diverse set of LLMs, both reasoning and non-reasoning models, on their ability to classify hierarchical ICD-10 codes from discharge summaries and evaluate the effect of structured reasoning on model performance.\\
\noindent\textit{Methods}: Using the MIMIC-IV dataset, the study selected 1,500 discharge summaries labeled with the top 10 most frequent ICD-10 codes, balancing dataset size with the high computational and financial cost of using LLMs. We first preprocessed the data to extract clinically relevant tokens before feeding it to the LLMs. Specifically, we used cTAKES, a clinical NLP tool, to identify medical concepts. Each summary was encoded and submitted to 11 LLMs using a standardized, structured prompt simulating a clinical coder. Models were evaluated using the F1 score across three ICD-10 levels for both primary and all diagnoses classification tasks.\\
\noindent\textit{Results}: The experimental results showed that none of the LLMs achieved F1 scores above 57\% for any of the classification tasks. Reasoning models on average outperformed non-reasoning models. The Gemini 2.5 Pro model demonstrated the highest performance across tasks. Performance declined as ICD code specificity increased. Certain codes, such as I25 (chronic ischemic heart disease), were classified more accurately than others, like Y92 or Z51, which consistently had near-zero F1 scores.\\
\noindent\textit{Conclusion}: LLMs, especially those equipped with structured reasoning capabilities, show promise in supporting clinical coders but fall short of autonomous deployment standards. Their effectiveness is constrained by ICD-10 granularity and variability across diagnostic categories. Future research should focus on hybrid models, domain-specific fine-tuning, and integration of structured clinical data.\\

\end{abstract}


\end{frontmatter}

\section{Introduction}

%\subsection{International Classification of Diseases}

\begin{figure}[h]
\centering
\includegraphics[width=0.7\textwidth]{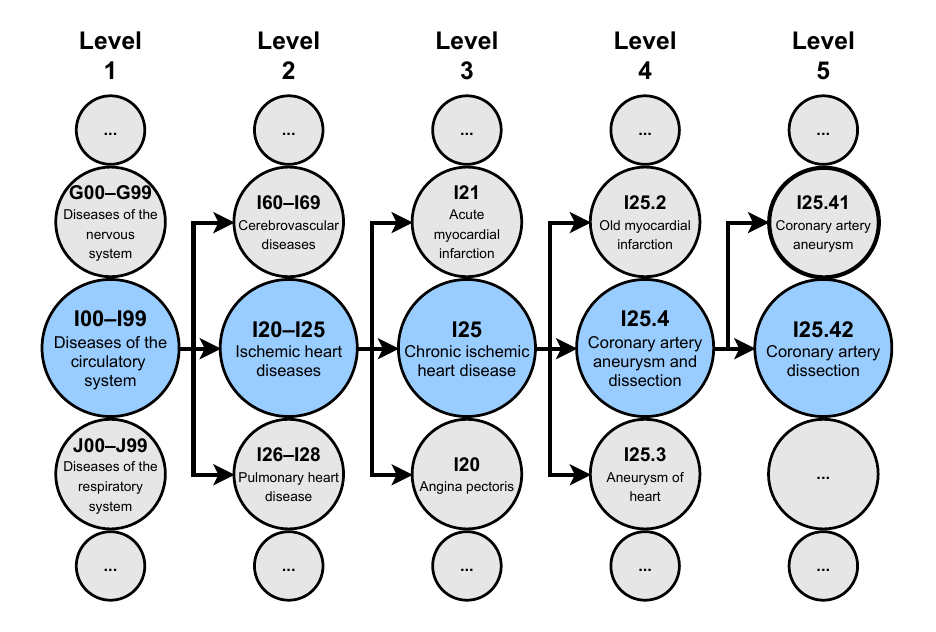}
\noindent
\caption{ICD-10 Code Hierarchy (Levels 1–5): This diagram illustrates the five-level structure of ICD-10 codes. The example hierarchy highlighted in blue is discussed in further detail in the text.}
\label{g_ICDHierarchy}
\end{figure}

The International Classification of Diseases, 10th Revision (ICD-10), provides a hierarchical framework for the systematic coding of diseases and health conditions. This structure comprises five distinct levels: chapters (Level 1), blocks (Level 2), categories (Level 3), subcategories (Level 4), and additional subdivisions (Level 5) \cite{992}. For instance, Chapter IX includes codes I00-I99, which pertain to ``Diseases of the circulatory system." Within this chapter, blocks such as I20-I25 classify ``Ischemic heart diseases." These blocks contain three-character categories (e.g., I25 for ``Chronic ischemic heart disease"), which may be further refined into four-character codes (e.g., I25.4 for ``Coronary artery aneurysm and dissection") and five-character codes (e.g., I25.42 for ``Coronary artery dissection"). An overview of this hierarchical structure is depicted in Figure \ref{g_ICDHierarchy}.

%\subsection{Clinical Coding}
Clinical coding transforms unstructured clinical documentation into structured ICD-10 codes. This process underpins accurate billing, health system planning, and research. Central to this task is the correct identification and classification of primary and secondary diagnoses. The primary diagnosis refers to the main condition prompting treatment, while secondary diagnoses capture comorbidities or complications that influence care \cite{993}.
Manual clinical coding remains the gold standard, with human coders achieving agreement rates of up to 87\% at Level 3 for principal diagnoses when standardized guidelines are applied \cite{994}. However, the process is time-consuming, prone to variability, and requires extensive training. Automated coding approaches have emerged to address these challenges, but they often fall short of human-level accuracy, especially in distinguishing primary versus secondary diagnoses, where context and clinical reasoning are essential \cite{996}.

%Leveraging this structure, researchers such as Mustafa \emph{et al.} \cite{922} have reframed clinical coding as a set of multi-label classification tasks across different levels of granularity. By training classifiers at specific levels (e.g., Level 2 across chapters or Level 3 within neoplasms), they reduce label complexity and improve accuracy by exploiting the ICD's inherent hierarchical organization.

%\subsection{Natural Language Processing}

To address the limitations of manual coding, natural language processing (NLP) technologies have been applied to extract structured information from unstructured texts such as discharge summaries. Tools like the Clinical Text Analysis and Knowledge Extraction System (cTAKES) \cite{470} have demonstrated strong performance in identifying clinical concepts through entity recognition. With task-specific terminologies and classification rules, cTAKES has been adapted for diagnostic validation, such as pneumonia identification from radiology reports, achieving accuracy of 76.3\% \cite{922, 999}.
Despite these advancements, automated ICD coding using NLP remains an active research area. Techniques involving machine learning and deep learning have been used to predict ICD codes, especially from discharge summaries \cite{813, 998}. 
Previous research \cite{922} has also reframed clinical coding as a set of multi-label classification tasks across different levels of granularity. By training classifiers at specific levels (e.g., Level 2 across chapters or Level 3 within neoplasms), label complexity is reduced and accuracy improved by exploiting the ICD's inherent hierarchical organization.
However, significant challenges remain, particularly the need for large, high-quality de-identified datasets and the difficulty of maintaining consistent accuracy across different clinical domains.

%\subsection{Large Language Models for Clinical Coding}

LLMs offer a promising direction for automating clinical coding. These models can process free-text clinical documentation and generate diagnostic codes with minimal human intervention. Studies show that domain-specific LLMs, especially those fine-tuned on biomedical corpora or augmented with ICD-10 guidelines, outperform general-purpose models in both summarizing clinical narratives and predicting diagnostic codes \cite{1000, 1001, 1002}. While general LLMs offer scalability, smaller, specialized models often yield better results in healthcare settings \cite{1003}.
Despite these improvements, limitations remain. LLMs may over-predict codes, and may misclassify or omit nuanced diagnoses \cite{1001}. These issues highlight the need for models that incorporate structured clinical reasoning to improve interpretability and diagnostic accuracy.

%\subsection{Reasoning in Large Language Models}

The effectiveness of LLMs in clinical settings increasingly depends on their ability to reason through complex information. Recent studies differentiate between ``reasoning LLMs" and ``non-reasoning LLMs" based on whether they employ explicit reasoning processes. Reasoning LLMs use structured prompts or chain-of-thought techniques to mimic clinical reasoning, thereby producing more accurate and interpretable outputs \cite{1005, 1006}. In contrast, non-reasoning LLMs generate answers directly, relying on statistical associations \cite{1007}.
Evidence suggests that structured reasoning significantly improves diagnostic performance. For instance, LLMs guided to summarize patient history before producing a diagnosis outperform those using standard prompts, with marked gains in accuracy \cite{1008}. Nevertheless, even reasoning-enhanced models struggle with more complex tasks such as treatment planning or differentiating subtle comorbidities \cite{1006, 1009, 985}.
A key advantage of reasoning LLMs is transparency: their step-by-step rationales enable clinicians to audit and trust the model’s decisions, addressing concerns about the “black-box” nature of AI systems \cite{1008}. As such, these models represent a crucial step toward integrating AI safely and effectively into clinical coding workflows.

%\section{The Problem}

LLMs have been developed with the ability to process and analyze clinical documents. Some of these models incorporate reasoning capabilities designed to enhance their performance in complex tasks such as medical coding \cite{981,1016, 1026}. This advancement raises several critical questions for the clinical informatics community.

First, can LLMs reach an acceptable accuracy when performing ICD-10 classification from unstructured clinical narratives? Unlike structured data, clinical notes often contain ambiguous, nuanced, and context-dependent information that requires domain expertise and inference. Evaluating LLMs in this setting is essential to understand whether they can meet the high standards of precision required for diagnostic coding.

Second, can LLMs serve effectively as intelligent assistants, augmenting the work of human coders? While full automation remains a long-term goal, many real-world clinical environments may benefit more from semi-automated workflows in which LLMs provide suggestions or confidence scores to support human decision-making.

Third, which LLMs are best suited for ICD-10 classification tasks, and does incorporating explicit reasoning capabilities in these models improve their performance compared to non-reasoning counterparts? Despite the growing number of LLMs available, there is limited systematic benchmarking of their capabilities, specifically in the medical coding domain and for hierarchical codes. Most comparative studies \cite{1023,1024,1025} focus on general NLP benchmarks rather than evaluating clinical coding accuracy or the impact of reasoning abilities on this complex task.

This paper aims to work toward addressing these questions by systematically evaluating the performance of selected LLMs on hierarchical ICD-10 code classification using discharge summaries.

\section{Materials and Methods}
\subsection{Dataset}
For our experiments, we utilized the MIMIC-IV database, which contains hundreds of thousands of high-quality, gold standard discharge summaries. From this dataset, we identified the top 10 most frequently occurring ICD-10 diagnosis codes across all discharge summaries. The selection was based on both primary and all diagnoses (including both primary and secondary types). These top 10 ICD-10 codes are listed in Tables \ref{table_Top10ICD}.
For each of the selected ICD-10 codes, we randomly sampled 150 discharge summaries, resulting in a total of 1,500 clinical documents for our study, balancing dataset size with the high computational and financial cost of using large language models. 
Each discharge summary in our dataset is multi-labeled, representing all associated ICD-10 diagnosis codes. This includes a single primary diagnosis code at each level. There can also be multiple secondary codes for each discharge summary.
%Each discharge summary in our dataset is multi-labeled, representing all associated hierarchical ICD-10 diagnosis codes (as shown in Fig.~\ref{g_ICDHierarchy}). This includes a single primary diagnosis code at each level. 

\begin{table}[t]

%\begin{table}
\caption{Top 10 ICD codes used in this study. The top part of the table shows 5 codes with the highest number of primary diagnoses cases, while the bottom part shows the top 5 codes with the maximum total cases in the MIMIC IV dataset.}
\label{table_Top10ICD}
\smallskip

\begin{tabular}{@{} L{190pt} C{40pt}C{40pt}C{85pt} @{}}
\toprule
			Diagnosis
			&\makecell{ICD-10\\Code}
                &\makecell{Total\\Cases}
			&\makecell{Primary\\Diagnosis Cases} \\

\midrule
		
			Sepsis	&A41	&7,430 &4,830	\\ 			
			Myocardial infarction	&I21	&5,735  &2,722	\\ 			
			Other medical care  &Z51	&6,919    &2,370	\\ 			
			Chronic ischaemic heart disease	&I25	&38,157  &2,302	\\ 			
			Hypertensive heart and renal disease    &I13 &8,366    &2,114	\\ 			
\midrule

			Disorders of lipoprotein metabolism and other lipidemias	&E78	&49,310 &9	\\ 			
			Essential hypertension	&I10	&43,574  &76	\\ 			
			Long term drug therapy	&Z79	&40,393  &0	\\ 						Personal history of other diseases and conditions    &Z87 &40,255    &0	\\ 			

			Place of occurrence of the external cause  &Y92	&35,297    &0	\\ 			
\bottomrule
\end{tabular}
%\end{table}

\end{table}

\subsection{Data Preparation}
To prepare the data for LLM input, discharge summaries were pre-processed using the cTAKES tool. cTAKES applies NLP and machine learning techniques to extract medically relevant tokens from unstructured clinical text. These tokens are classified into categories such as medications, symptoms, diseases, laboratory results, and procedures. Additionally, cTAKES identifies the polarity of each term, indicating whether the clinical concept is affirmed or negated in the narrative.
Following extraction, each discharge summary was reconstructed to include only the identified medical terminologies. For each term, its frequency of occurrence and polarity (affirmed or negated) were preserved. This token-based representation of the text ensures that only clinically meaningful content is retained.

This approach offers multiple benefits. First, it removes non-essential words (e.g., stop words), allowing LLMs to focus exclusively on medically relevant information. This not only enhances the clinical interpretability of the prompt but also aligns the model’s attention with diagnostic cues. Second, it significantly reduces the prompt size, which is particularly important due to the input length limitations of most LLMs. By condensing the input to only the most important clinical content, the processing becomes more efficient. Finally, the reduction in token count translates into lower computational costs, as many LLM services charge based on the number of tokens processed.
%Overall, this structured preprocessing pipeline improves model performance, optimizes resource use, and maintains the clinical integrity of the original discharge summaries.

\subsection{LLMs}

In our experiments, we selected a diverse set of state-of-the-art LLMs, including both reasoning and non-reasoning variants, to evaluate their performance in clinical coding tasks. The reasoning models included DeepSeek Reasoner, Gemini 2.0 Flash Thinking, Gemini 2.5 Pro, GPT o3 Mini, and Qwen QWQ, which are designed to follow structured, multi-step reasoning processes that enhance interpretability and diagnostic accuracy \cite{1013, 1014}. In contrast, the non-reasoning models, Llama 4 Scout, GPT 4o, GPT 4o Mini, Llama 3.3 Versatile, DeepSeek Chat, and Gemini 2.0 Flash, are optimized for efficiency and direct response generation without explicit reasoning steps \cite{1015}. These models were accessed through various platforms depending on availability; some were used via their official providers, while others were accessed through third-party hosting platforms. Table \ref{table_LLMPlatform} details the models used and their corresponding access platforms. This diverse selection of LLMs allowed us to compare the diagnostic coding accuracy and behavior across reasoning vs. non-reasoning paradigms.

\begin{table}[!t]
\caption{Large Languages Models used.}
\label{table_LLMPlatform}
\begin{tabular}{@{}L{180pt}C{100pt}C{90pt} @{}}
\toprule

\textbf{\makecell{Model}}
&\textbf{\makecell{Reasoning Type}}
&\textbf{\makecell{Platform}} \\

\midrule
		
Deepseek Reasoner	        &Reasoning	&Deepseek \\
Gemini 2.0 Flash Thinking	&Reasoning	&Google \\
Gemini 2.5 Pro	            &Reasoning	&Google \\
GPT o3 Mini                 &Reasoning	&OpenAI \\
Qwen QWQ                &Reasoning	&Groq \\
\midrule
Deepseek Chat   	        &Non-Reasoning	&Deepseek \\
Gemini 2.0 Flash        	&Non-Reasoning	&Google \\
GPT 4o                  &Non-Reasoning	&OpenAI \\
GPT 4o Mini                 &Non-Reasoning	&OpenAI \\
Llama 3.3 Versatile     &Non-Reasoning	&Groq \\
Llama 4 Scout     &Non-Reasoning	&Groq \\

\bottomrule
\end{tabular}

\end{table}
\subsection{Prompt Engineering}
To guide the LLMs in performing ICD-10 code classification, a structured prompt was developed to simulate the role of a Clinical Coder. The prompt begins with an explicit instruction, “In the role of Clinical Coder” to frame the model's response contextually. It then presents the tokenized discharge summary text under the placeholder [RepText] and asks two specific questions: identifying the primary ICD-10 code and listing any secondary codes. To ensure consistency and facilitate automated evaluation, the expected response format was standardized using labeled sections: “\#Primary\#” followed by the answer to the first question, and “\#Secondary\#” for the second. Furthermore, the prompt instructs the model to provide answers “as simple as possible” and in the precise ICD-10 format (\#\#\#.\#\#\#), reducing ambiguity and aligning with coding standards. The full prompt used in this study is as follows: \\\\
\noindent\fbox{%
    \parbox{\textwidth}{%

\textit{“In the role of Clinical Coder, analyze the following medical reports and answer the questions in the specified format
[RepText]\\
Provide the answers directly as simple as possible and in this format:\\
\#Primary\# Then the answer for the first question: What is the primary ICD-10 code diagnosis in format of \#\#\#.\#\#\# ?\\
\#Secondary\# Then the answer for the second question: What are the secondary ICD-10 Code diagnoses in format of \#\#\#.\#\#\# ?”}
}
}

\subsection{Evaluation Metrics}

We adopted the F1 score as the primary evaluation metric due to its ability to effectively balance precision and recall in multi-label classification tasks \cite{783}. This metric provides a holistic assessment of model performance by aggregating results across all discharge summaries. Specifically, the F1 score is sensitive to the model’s ability to correctly identify frequently occurring codes, making it particularly suitable for the imbalanced nature of clinical coding tasks.
The F1 score was computed using its standard formulation, which requires identifying true positives (TP), false positives (FP), and false negatives (FN), as defined in Equation~(\ref{eq:f1}):

\begin{equation}
F_1 = \frac{2 \cdot TP}{2 \cdot TP + FP + FN}.
\label{eq:f1}
\end{equation}

% \begin{figure}[h]
% \centering
% \includegraphics[width=0.7\textwidth]{graph_F1Venn.pdf}
% \caption{The relationship among actual, predicted, and matched codes used in F1 score computation.}
% \label{g_F1Venn}
% \end{figure}

% Figure~\ref{g_F1Venn} illustrates the relationships among matched codes (true positives, represented in red), predicted codes that do not match actual codes (false positives, in blue), and actual codes not identified by the model (false negatives, in green). The F1 score calculation involves aggregating these counts across all discharge summaries prior to computing the overall score, as shown in Equations~(\ref{eq:f1_s1}) and~(\ref{eq:f1_s2}):

% \begin{equation}
% F_1 = \frac{2 \cdot TP}{(FP + TP) + (FN + TP)}
% \label{eq:f1_s1}
% \end{equation}

% \begin{equation}
% F_1 = \frac{2 \cdot \sum |\text{Matching Codes}|}{\sum |\text{Actual Codes}| + \sum |\text{Predicted Codes}|}
% \label{eq:f1_s2}
% \end{equation}

To ensure a comprehensive evaluation of model performance, we calculated F1 scores across six distinct settings. These included two clinical coding scenarios, all diagnoses (including both primary and secondary diagnoses) and primary diagnoses only, combined with three levels of ICD-10 code granularity: Level 5 (most specific; e.g., I25.42), Level 4 (e.g., I25.4), and Level 3 (more general; e.g., I25). This structure enabled a detailed analysis of each model’s capability to assign both specific and general diagnostic codes accurately.

\subsection{Methodology}

\begin{figure}[H]
\centering
\includegraphics[width=1.0\textwidth]{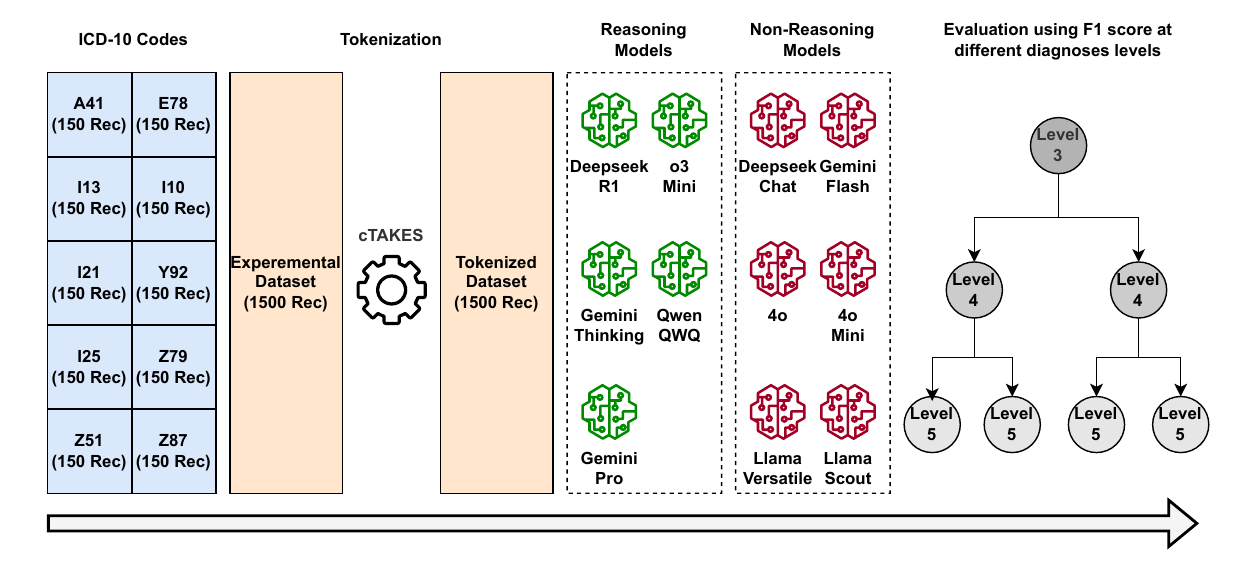}
\noindent
\caption{The Workflow used for submitting discharge summaries to LLMs, extracting ICD-10 codes, and evaluating performance using F1 scores across ICD-10 levels (Level 3, 4, and 5).}
\label{g_Process}
\end{figure}

The tokenized dataset was submitted to the various LLM platforms one record at a time using API calls. To automate this procedure, a custom desktop application was developed to handle the submission of each discharge summary. This application called each model’s API 1,500 times, corresponding to the 1,500 discharge summaries used in the experiment, resulting in a total of 16,500 API calls across all 11 models. The output from the models was categorized into primary and secondary results. The primary output contained a single, identified ICD-10 code, whereas the Secondary output included a list of ICD-10 codes that were often embedded within additional contextual text. Both outputs were parsed to extract ICD-10 codes and combined to form a comprehensive list of predicted diagnoses per summary. These predictions were then compared to the actual codes to evaluate model performance. Specifically, the number of matching ICD-10 codes, actual codes, and predicted codes were used to calculate F1 score at the most granular level (Level 5) ICD-10 codes, and subsequently at Levels 4 and 3 to assess hierarchical performance. The complete process is visually summarized in Figure \ref{g_Process}.

\section{Results}

A total of 16,500 classification outcomes were evaluated to assess the performance of various LLMs in medical code classification tasks. While LLMs have demonstrated notable advancements in NLP, the results revealed considerable limitations in their current ability to accurately assign ICD-10 codes. Notably, no model or configuration achieved an F1 score exceeding 56\%, underscoring the immaturity of existing LLMs for high precision clinical code classification.

\subsection{Primary diagnosis classification}

Our analysis revealed that the classification performance of ICD-10 codes by LLMs can be broadly grouped into three distinct categories. First, certain ICD-10 codes consistently yielded exceptionally low performance in the primary diagnosis classification task. Specifically, codes I13, Y92, and Z51 exhibited near-zero F1 scores across all three granularity levels and across nearly all evaluated models. For I13 and Y92, most models achieved an F1 score of 0\%, with none surpassing 2\% for I13 and 5\% for Y92. Similarly, Z51 produced a uniform F1 score of 0\% across all levels and models, highlighting persistent challenges in capturing these diagnostic categories. These findings suggest that some codes are fundamentally more difficult for LLMs to detect as a primary diagnosis. Tables~\ref{table_I13},~\ref{table_Y92}, and~\ref{table_Z51} in the appendix present these results in detail.

Second, investigation showed that performance degradation became more pronounced at the most granular classification level (Level 5), particularly for codes such as A41, I21, and Z87, as shown in Appendix Tables~\ref{table_A41},~\ref{table_I21}, and~\ref{table_Z87}. While moderate accuracy was observed at Level 3 for some models, performance sharply declined at higher levels. For example, A41 was only moderately identified by Gemini 2.5 Pro and Qwen QWQ, which attained close to 15\% F1 score at Level 3, whereas all other models scored below 4\% at primary Levels 4 and 5. I21 presented an atypical case: DeepSeek Chat achieved an F1 score of 41\% at Level 3 but completely failed at Levels 4 and 5, indicating close yet non-specific predictions. For Z87, most models, including GPT o3 Mini and LLaMA 3.3 Versatile, scored very low, mostly 0\% across all levels, but Gemini 2.5 Pro, marginally improved to 26\% at Level 3.

Third, as shown in Appendix Table \ref{table_E78}, moderate results were observed for E78, where DeepSeek Chat, DeepSeek Reasoner, and Gemini 2.5 Pro each attained an F1 score of 21\% at Level 3. GPT o3 Mini demonstrated relatively better results, achieving F1 scores of 29\%, 22\%, and 12\% at Levels 3, 4, and 5, respectively. Similarly, for I10, GPT o3 Mini outperformed others at Levels 3 and 4, achieving F1 scores of 26\% and 23\%, respectively. At Level 5, Gemini 2.5 Pro achieved the highest score for this code at 11\%. Appendix Table~\ref{table_I10} presents these results in detail.

The strongest performance across all codes was observed for I25, which achieved the highest average F1 score across all ICD-10 codes. Seven models exceeded a 30\% F1 score at Level 3 primary, with GPT o3 Mini outperforming all others, reaching over 54\% at both Levels 3 and 4. Even though it was not the leader at level 5, it maintained a score of 22\% at this level, where overall performance typically drops below 12\% for all other ICD-10 codes. I25 was also unique in that the primary diagnosis classification surpassed performance on all diagnosis-level tasks. Appendix Table~\ref{table_I25} presents these results.
In contrast, Appendix Table~\ref{table_Z79} shows that Z79 yielded relatively weak results. DeepSeek Reasoner achieved an F1 score of 20\% at Level 3 and declined to 8\% at Level 5. At Level 4, both DeepSeek Reasoner and Gemini 2.5 Pro tied with a modest performance of 12\%.

%%%%%%%%%%%%%%%%%%%%%%%%%%%
\subsection{All diagnosis classification}
In the all diagnoses classification task, model performance improved notably compared to the primary diagnosis task. This broader scope appeared to align more effectively with the strengths of LLMs, as it imposed less penalization for predictions that were close but not exact. Among all tested models, Gemini 2.5 Pro emerged as the most consistent and highest performing across nearly all ICD-10 codes and levels. Refer to Appendix Tables~\ref{table_A41} to~\ref{table_Z87} for detailed results. I25 emerged as the strongest performing code overall, with F1 scores of 46\% at Level 3, 35\% at Level 4, and 30\% at Level 5, indicating the Gemini 2.5 Pro’s robustness even at the most granular level of classification.
The only exception to Gemini 2.5 Pro’s dominance was observed with the code Y92. Although it led at Level 3 with an F1 score of 26\%, its advantage over DeepSeek Reasoner was marginal, amounting to less than 0.15\%. At Levels 4 and 5, DeepSeek Reasoner surpassed Gemini 2.5 Pro, achieving F1 scores of 16\% and 12\%, respectively. These results collectively suggest that broader diagnostic classification is currently better aligned with the capabilities of LLMs and that performance is highly dependent on both the specific ICD-10 code and the level of classification granularity.

\subsection{Comparison}
%%%%%%%%%%%%%%%%%%%%%%%%%%%
The comparison between the primary diagnosis and all diagnoses classification tasks further highlights the divergence in model performance under varying levels of diagnostic complexity. Results consistently favored the all diagnoses task, a trend that is conceptually intuitive. This task allows the assignment of multiple ICD-10 codes, including both primary and secondary diagnoses, providing greater flexibility and enabling models to reflect partial correctness. In contrast, the primary diagnosis task demands the identification of a single, most accurate code, making it inherently more restrictive and challenging. This distinction had a notable impact on overall performance.
Across both tasks, a consistent decline in accuracy was observed as the level of code granularity increased. Models performed best at Level 3, followed by Level 4, with Level 5 yielding the lowest accuracy scores. This downward trend aligns with the increasing specificity and semantic complexity of deeper ICD-10 classifications. The patterns are further illustrated in Appendix Tables~\ref{table_A41} through~\ref{table_Z87} and Appendix Charts~\ref{MiPriA41} to~\ref{MiAllZ87}.
%%%%%%%%%%%%%%%%%%%%%%%%%%%%%%%%%

When examining average F1 scores across all ICD-10 codes, Gemini 2.5 Pro emerged as the most effective model. At Level 3, it achieved an F1 score of 16\% in the primary diagnosis task and 39\% in the all diagnoses task, representing the highest performance recorded at that level. Conversely, Llama 4 Scout demonstrated the weakest results at Level 3 in the primary diagnosis task, with an F1 score of only 5\%, while at 14\%, Gemini 2.0 Flash Thinking produced the lowest score for the all diagnoses task at the same level. Refer to Table~\ref{table_Avg}.

\begin{table}[h]
    \caption{Average F1 Scores for Primary and All Diagnoses Across ICD-10 Levels 3 to 5 for all LLMs}
    \label{table_Avg}
    \begin{tabular}{lllllll}
    \toprule
        Model  & \makecell{Level 3\\Primary} & \makecell{Level 4\\Primary} &  \makecell{Level 5\\Primary} & \makecell{Level 3\\All} &  \makecell{Level 4\\All} & \makecell{Level 5\\All}  \\
    \midrule
Deepseek Reasoner &13.3\% &10.1\% &4.5\% &31.4\% &20.0\% &16.7\% \\
Gemini 2.0 Flash\\Thinking &9.5\% &7.9\% &1.1\% &13.8\% &8.3\% &4.3\% \\
Gemini 2.5 Pro &\textbf{16.3\%} &\textbf{13.0\%} &\textbf{7.5\%} &\textbf{39.1\%} &\textbf{27.4\%} &\textbf{23.7\%} \\
GPT o3 Mini &12.4\% &10.9\% &4.4\% &21.3\% &12.2\% &7.6\% \\
Qwen QWQ &10.0\% &4.9\% &0.7\% &22.4\% &10.0\% &3.8\% \\
    \midrule

Deepseek Chat &11.6\% &3.3\% &1.5\% &28.2\% &15.6\% &11.9\% \\
Gemini 2.0 Flash &6.5\% &5.8\% &2.1\% &17.4\% &9.6\% &7.9\% \\
GPT 4o &8.7\% &7.6\% &2.8\% &21.5\% &12.7\% &10.2\% \\
GPT 4o Mini &8.6\% &6.9\% &2.9\% &24.4\% &13.2\% &10.7\% \\
Llama 3.3 Versatile &8.8\% &5.4\% &2.4\% &20.8\% &10.3\% &8.3\% \\
Llama 4 Scout &4.6\% &0.6\% &0.1\% &17.3\% &6.5\% &4.5\% \\

  \bottomrule
\end{tabular}
    \caption*{\textbf{Note:} Bold values indicate the highest F1 score in each column.}
\end{table}

This performance hierarchy was maintained at Level 4, where Gemini 2.5 Pro attained an F1 score of 13\% in the primary diagnosis task and 27\% in the all diagnoses task. Once again, Llama 4 Scout remained the lowest performing model, registering F1 scores of 1\% and 7\%, respectively. At Level 5, the most demanding level of diagnostic precision, Gemini 2.5 Pro sustained its lead with F1 scores of 8\% for primary diagnosis and 24\% for all diagnoses. In contrast, Llama 4 Scout recorded an exceptionally low score of just 0.1\% in the primary diagnosis task, while Qwen QWQ exhibited the weakest performance in the all diagnoses task at this level, managing only 4\%. These outcomes underscore the compounded difficulty that models face at finer levels of diagnostic specificity and reinforce the relative robustness of Gemini 2.5 Pro across both classification scenarios.

%%%%%%%%%%%%%%%%%%%%%%%%%%%%%%%%%%%

\begin{figure}[H]
\centering
\includegraphics[width=1.0\textwidth]{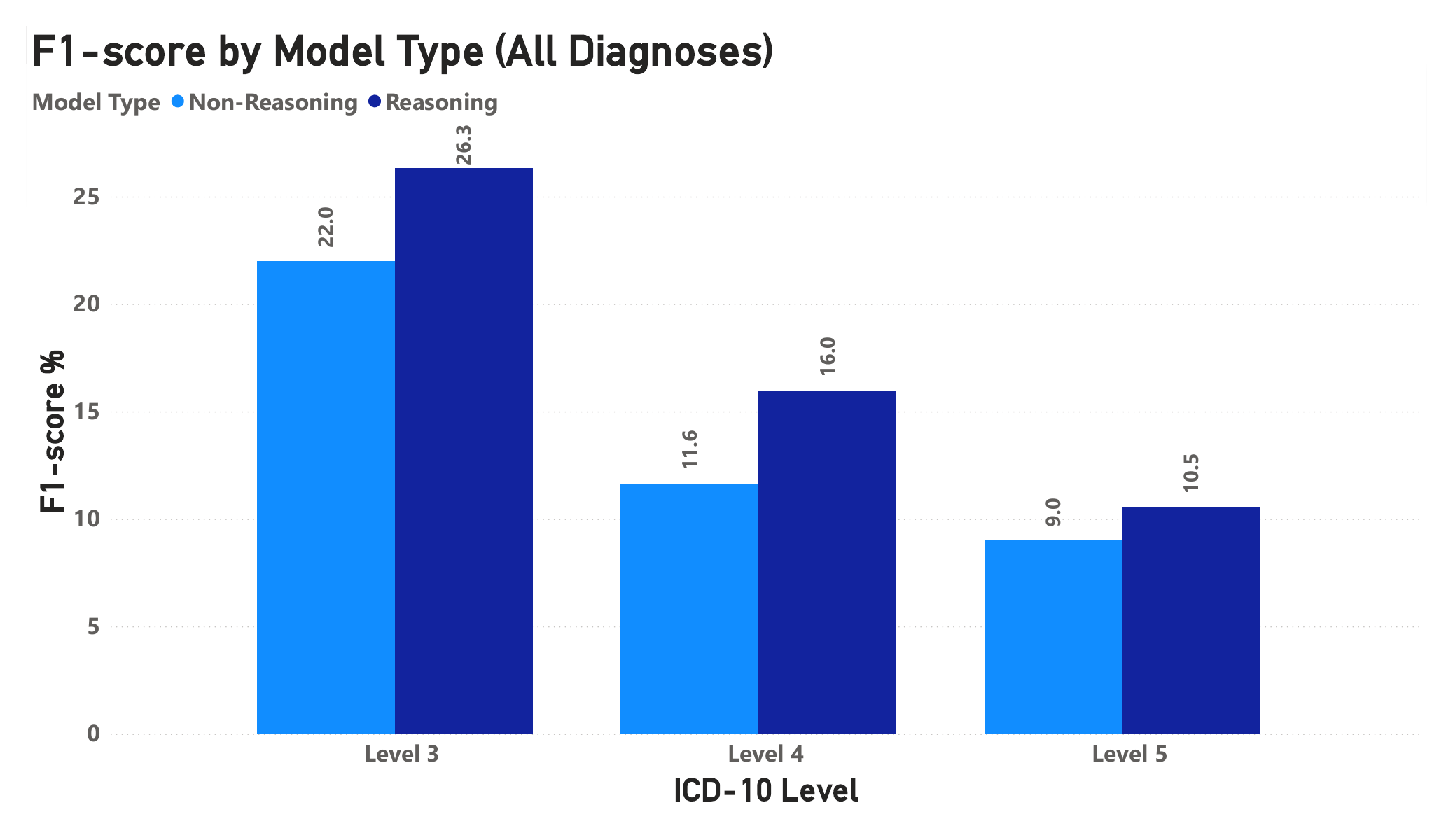}
\noindent
\caption{Comparison of average F1 Scores by Model Reasoning Type (All Diagnoses).}
\label{MiReasonAll}
\end{figure}

\begin{figure}[H]
\centering
\includegraphics[width=1.0\textwidth]{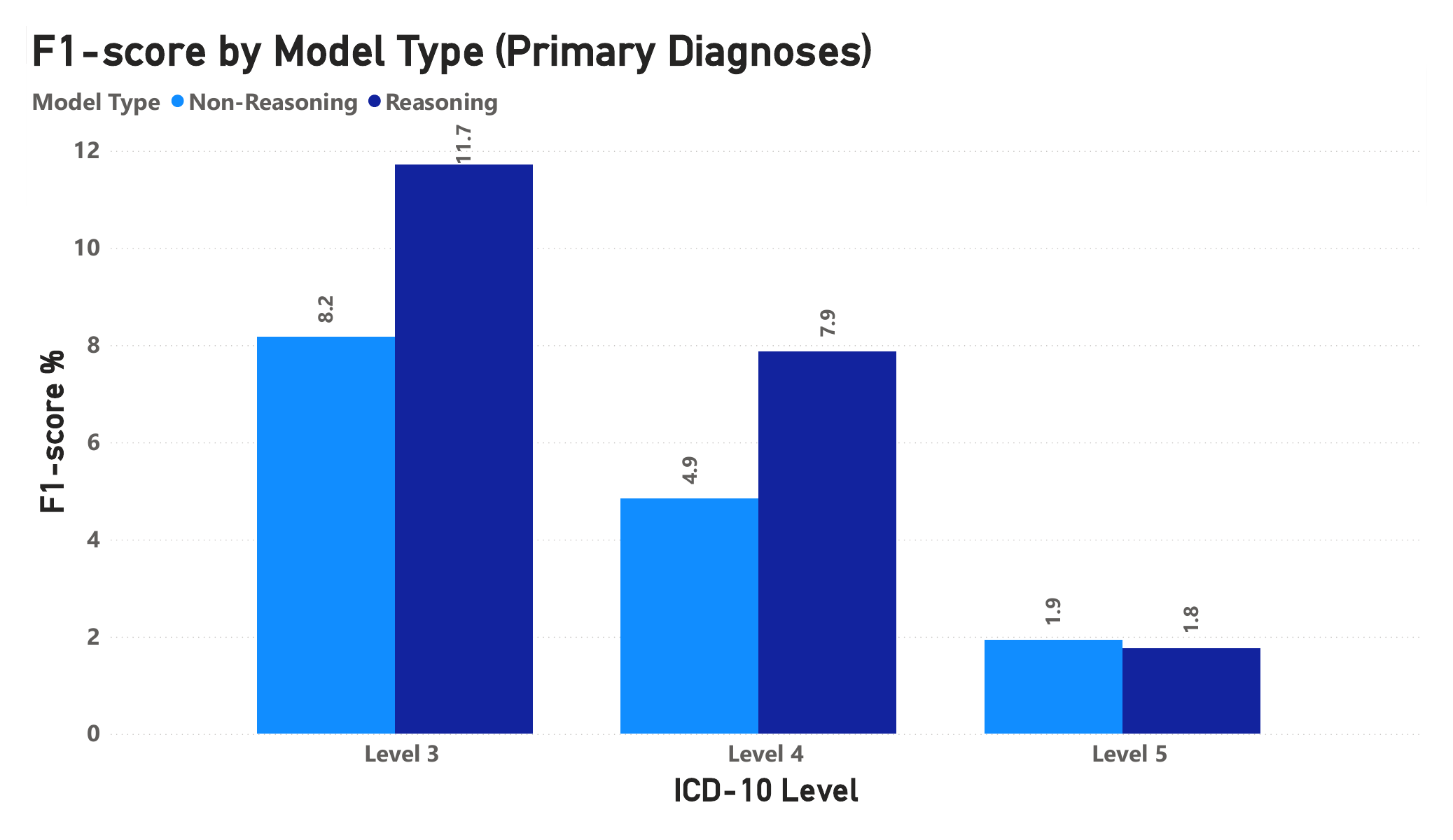}
\noindent
\caption{Comparison of F1 Scores by Model Reasoning Type (Primary Diagnoses).}
\label{MiReasonPri}
\end{figure}

Across almost all ICD-10 levels and diagnosis types, reasoning models on average outperform non-reasoning models, indicating that reasoning capabilities enhance a model's ability to classify accurately, as reflected in higher F1 scores. For example, at Level 3 within the all diagnoses results, reasoning models achieved an F1 score of 26\%, compared to 22\% for non-reasoning models (see Figure \ref{MiReasonAll}). This performance gap persists across different levels and diagnosis categories. Additionally, F1 scores generally decline as ICD-10 levels become more granular, from Level 3 to Level 5. Within the all diagnoses group, both model types follow this downward trend. Moreover, as shown in Figure \ref{MiReasonPri}, models evaluated on primary diagnosis consistently achieve lower F1 scores than those evaluated on all diagnoses. For instance, the reasoning model’s performance drops from 26\% on all diagnoses to 12\% on primary diagnosis at Level 3. In the level 5 classification case for primary diagnoses, non-reasoning models are slightly better than reasoning ones, but this is very minor and happens at the most complex granular level 5.

\section{Discussion}

This study presents a systematic evaluation of LLMs for ICD-10 code classification from clinical narratives, with particular emphasis on the role of reasoning capabilities in enhancing performance. Our findings offer several key insights into the practical utility and current limitations of LLMs in the context of medical coding.
On average, reasoning augmented LLMs outperformed non-reasoning models, especially in tasks requiring semantic interpretation of complex clinical text. Notably, models such as Gemini 2.5 Pro and DeepSeek Reasoner demonstrated higher performance across multiple granularity levels and diagnosis types, highlighting the value of incorporating intermediate reasoning steps that mimic human cognitive processes. These results suggest that LLMs with advanced reasoning capabilities may better interpret clinical narratives by identifying implicit associations among symptoms, test results, and medical procedures.

An interesting trend observed was the improved accuracy at higher ICD-10 hierarchy levels. While detailed classification at the most granular code level remains a challenge, performance at broader categorical levels was significantly more reliable. This finding carries important implications for real-world applications, where general diagnostic classification can still provide meaningful clinical and operational value.

However, model performance was not uniformly strong across all ICD-10 code categories. Certain conditions, such as chronic ischemic heart disease (I25), were more accurately classified than others, particularly aftercare and medical care codes. This inconsistency underscores the importance of designing model prompts and training strategies tailored to the linguistic and clinical complexity of specific domains.

A significant contribution of this study is the novel integration of cTAKES with LLMs for ICD-10 classification, a combination not previously explored in the literature. By applying cTAKES to extract structured clinical concepts from real discharge summaries in the MIMIC-IV dataset, we enhanced the contextual grounding of the input to LLMs. This approach enabled a more realistic and clinically meaningful evaluation of LLM capabilities, closely reflecting the complexity of actual hospital coding workflows.

While the evaluation focused on a primary set of 10 ICD-10 codes for targeted analysis, the model was open to predicting any secondary codes across the full ICD-10 spectrum. This broader scope allowed the LLMs to demonstrate their generalizability beyond the predefined categories, capturing a wider range of clinically relevant diagnoses present in the discharge summaries. This approach provides added value by better reflecting real-world coding scenarios, where both primary and secondary conditions contribute to patient care and reimbursement. However, the primary code selection and the limited sample size of 1,500 discharge summaries may still introduce selection bias and affect generalizability. 

Additionally, while reasoning models showed clear advantages, these were most pronounced at higher abstraction levels, with only modest improvements at the most detailed coding level. This underscores the need for continued refinement through domain-specific fine-tuning, expansion to more diverse and comprehensive code sets, and the development of hybrid systems that integrate rule-based logic with LLMs to enhance both precision and coding granularity.

Moreover, while LLMs show potential as assistive tools for clinical coders, they are not yet suitable for fully autonomous deployment. Current models remain vulnerable to hallucinations and lack access to structured data such as laboratory results or vital signs, which are often critical for accurate coding decisions and are used by clinical coders. 

Future research should explore multimodal models that integrate structured and unstructured data, leverage few-shot and in-context learning tailored to medical text, and assess ensemble strategies that combine the strengths of multiple LLMs. Lastly, the ethical and regulatory implications of deploying LLMs in healthcare settings must not be overlooked. Issues such as transparency, auditability, and compliance with health information privacy laws are fundamental prerequisites for the safe and effective integration of these technologies into clinical practice.

\section{Conclusion}
This study provides evidence that LLMs enhanced with reasoning capabilities can outperform standard models in automated ICD-10 classification tasks. Models like DeepSeek Reasoner and Gemini 2.5 Pro consistently delivered higher F1 scores across both primary and all diagnoses tasks and at multiple levels of ICD-10 granularity. Their relative strength at broader classification levels indicates a better ability to comprehend complex clinical narratives and accurately infer diagnostic categories.
The observed decline in performance at more granular coding levels underscores the challenge of achieving fine-level diagnostic precision. However, the performance advantage of reasoning models suggests they hold significant promise as supportive tools for semi-automated clinical coding, particularly for higher level classifications.

These findings have important implications for healthcare automation. Integrating reasoning-capable LLMs into clinical coding workflows could enhance coding efficiency, reduce human workload, and improve coding accuracy. For developers and researchers in clinical NLP, this emphasizes the need to prioritize reasoning strategies when selecting or training LLMs for healthcare applications. Future work should focus on domain-specific fine-tuning, expanding datasets with diverse clinical conditions, and combining multiple model outputs through ensemble approaches. These advancements will be essential to overcoming current limitations and progressing toward fully automated, high-precision clinical coding solutions.

\section*{Declarations}

\subsection*{Competing Interests}
The authors declare that they have no known competing financial interests or personal relationships that could have appeared to influence the work reported in this paper.

\subsection*{Ethical Compliance}
This study was conducted using publicly available, de-identified data from the MIMIC-IV database, which is approved for research use. No additional ethical approval was required as no direct patient contact or identifiable patient information was involved.

\subsection*{Informed Consent}
Not applicable. This study did not involve human participants, animals, or identifiable personal data.

\subsection*{Funding Sources}
This research did not receive any specific grant from funding agencies in the public, commercial, or not-for-profit sectors.

\subsection*{Large Language Model Usage Disclosure}
Large Language Models such as ChatGPT were used for language editing and improving the clarity of the manuscript. No LLMs were used for generating scientific content or interpreting study results.

\clearpage

\bibliographystyle{elsarticle-num} 
\bibliography{ref.bib}% common bib file

\clearpage

\begin{appendices}
\section{Appendix}
\subsection{Tables}

\begin{table}[h]
    \caption{F1 Scores for A41 Primary and All Diagnoses (Levels 3-5) by LLM}
    \label{table_A41}
    \begin{tabular}{lllllll}
    \toprule
        Model  & \makecell{Level 3\\Primary} & \makecell{Level 4\\Primary} &  \makecell{Level 5\\Primary} & \makecell{Level 3\\All} &  \makecell{Level 4\\All} & \makecell{Level 5\\All}  \\
    \midrule
      	
Deepseek Reasoner &5.9\% &3.9\% &3.3\% &27.4\% &17.4\% &16.0\% \\
Gemini 2.0 Flash\\Thinking &2.0\% &1.3\% &1.0\% &10.4\% &5.9\% &4.1\% \\
Gemini 2.5 Pro &14.7\% &\textbf{8.0}\% &\textbf{8.0}\% &\textbf{39.3}\% &\textbf{26.7}\% &\textbf{23.8}\% \\
GPT o3 Mini &2.3\% &1.3\% &0.7\% &18.3\% &11.4\% &5.5\% \\
Qwen QWQ &\textbf{14.9}\% &5.2\% &1.1\% &19\% &8.3\% &3.2\% \\
\midrule
Deepseek Chat &2.6\% &1.3\% &1.3\% &27.1\% &12.9\% &10.4\% \\
Gemini 2.0 Flash &0.0\% &0.0\% &0.0\% &15.5\% &7.9\% &6.6\% \\
GPT 4o &2.7\% &1.9\% &0.7\% &19.8\% &11.3\% &9.2\% \\
GPT 4o Mini &1.9\% &1.3\% &1.0\% &25.2\% &11.9\% &10.4\% \\
Llama 3.3 Versatile &2.1\% &1.0\% &1.0\% &17.1\% &8.6\% &7.5\% \\
Llama 4 Scout &4.4\% &0.7\% &0.0\% &14.2\% &5.8\% &3.8\% \\

  \bottomrule
\end{tabular}
    \caption*{\textbf{Note:} Bold values indicate the highest F1 score in each column.}
\end{table}

\begin{table}[h]
    \caption{F1 Scores for E78 Primary and All Diagnoses (Levels 3-5) by LLM}
    \label{table_E78}
    \begin{tabular}{lllllll}
    \toprule
        Model  & \makecell{Level 3\\Primary} & \makecell{Level 4\\Primary} &  \makecell{Level 5\\Primary} & \makecell{Level 3\\All} &  \makecell{Level 4\\All} & \makecell{Level 5\\All}  \\
    \midrule
Deepseek Reasoner &21.3\% &16.3\% &6.6\% &31.2\% &19.9\% &17.7\% \\
Gemini 2.0 Flash\\Thinking &11.7\% &8.6\% &1.1\% &12.1\% &6.6\% &3.9\% \\
Gemini 2.5 Pro &21.3\% &17.3\% &11.3\% &\textbf{40.3\%} &\textbf{28.4\%} &\textbf{26.3\%} \\
GPT o3 Mini &\textbf{29.3\%} &\textbf{22.3\%} &\textbf{12.0\%} &19.1\% &11.2\% &6.5\% \\
Qwen QWQ &12.5\% &4.3\% &0.9\% &20.8\% &9.4\% &3.4\% \\
    \midrule
Deepseek Chat &21.1\% &5.2\% &3.5\% &28.1\% &15.1\% &12.3\% \\
Gemini 2.0 Flash &10.6\% &9.4\% &3.2\% &18.1\% &9.6\% &7.2\% \\
GPT 4o &12.7\% &10.7\% &6.0\% &22.1\% &12.7\% &11\% \\
GPT 4o Mini &3.6\% &1.4\% &1.4\% &25.4\% &12.8\% &10.9\% \\
Llama 3.3 Versatile &18.1\% &10.0\% &5.5\% &21.7\% &10.5\% &8.8\% \\
Llama 4 Scout &8.0\% &1.0\% &0.0\% &18.5\% &6.7\% &4.7\% \\

  \bottomrule
\end{tabular}
    \caption*{\textbf{Note:} Bold values indicate the highest F1 score in each column.}

\end{table}

\begin{table}[h]
    \caption{F1 Scores for I10 Primary and All Diagnoses (Levels 3-5) by LLM}
    \label{table_I10}
    \begin{tabular}{lllllll}
    \toprule
        Model  & \makecell{Level 3\\Primary} & \makecell{Level 4\\Primary} &  \makecell{Level 5\\Primary} & \makecell{Level 3\\All} &  \makecell{Level 4\\All} & \makecell{Level 5\\All}  \\
    \midrule

Deepseek Reasoner &23.6\% &18.3\% &6.9\% &32.1\% &23.8\% &19.6\% \\
Gemini 2.0 Flash\\Thinking &16.0\% &14.3\% &3.0\% &14.9\% &9.6\% &5.8\% \\
Gemini 2.5 Pro &24.2\% &21.5\% &\textbf{11.4\%} &\textbf{41.6\%} &\textbf{31.6\%} &\textbf{28.8\%} \\
GPT o3 Mini &\textbf{26\%} &\textbf{22.7\%} &9.0\% &19.4\% &10.9\% &8.2\% \\
Qwen QWQ &12.5\% &6.1\% &0.7\% &21.7\% &11.8\% &4.6\% \\
    \midrule

Deepseek Chat &20.3\% &7.8\% &2.1\% &26.1\% &14.4\% &10.8\% \\
Gemini 2.0 Flash &10.4\% &9.2\% &3.5\% &20.4\% &13.3\% &10.4\% \\
GPT 4o &8.7\% &8.3\% &4.7\% &22.7\% &15.2\% &13.0\% \\
GPT 4o Mini &20.7\% &16.1\% &8.9\% &23.9\% &14.7\% &12.8\% \\
Llama 3.3 Versatile &19.2\% &9.4\% &4.5\% &21.2\% &12.0\% &9.5\% \\
Llama 4 Scout &9.6\% &0.8\% &0.4\% &17.9\% &9.3\% &6.4\% \\

  \bottomrule
\end{tabular}
    \caption*{\textbf{Note:} Bold values indicate the highest F1 score in each column.}

\end{table}

\begin{table}[h]
    \caption{F1 Scores for I13 Primary and All Diagnoses (Levels 3-5) by LLM}
    \label{table_I13}
    \begin{tabular}{lllllll}
    \toprule
        Model  & \makecell{Level 3\\Primary} & \makecell{Level 4\\Primary} &  \makecell{Level 5\\Primary} & \makecell{Level 3\\All} &  \makecell{Level 4\\All} & \makecell{Level 5\\All}  \\
    \midrule

Deepseek Reasoner &1.3\% &0.0\% &0.0\% &35.3\% &15.7\% &14.7\% \\
Gemini 2.0 Flash\\Thinking &0.0\% &0.0\% &0.0\% &14.1\% &5.9\% &3.7\% \\
Gemini 2.5 Pro &0.7\% &0.0\% &0.0\% &\textbf{42.7\%} &\textbf{24.3\%} &\textbf{22.3\%} \\
GPT o3 Mini &0.0\% &0.0\% &0.0\% &24.9\% &9.2\% &6.0\% \\
Qwen QWQ &0.5\% &0.2\% &0.1\% &25.3\% &8.6\% &2.9\% \\
    \midrule
Deepseek Chat &0.0\% &0.0\% &0.0\% &33.4\% &16.3\% &11.8\% \\
Gemini 2.0 Flash &0.0\% &0.0\% &0.0\% &17.6\% &7.2\% &5.5\% \\
GPT 4o &0.0\% &0.0\% &0.0\% &24.4\% &10.6\% &8.7\% \\
GPT 4o Mini &0.6\% &\textbf{0.6\%} &\textbf{0.6\%} &31.3\% &12.8\% &11.2\% \\
Llama 3.3 Versatile &1.0\% &0.5\% &0.5\% &25.8\% &10.4\% &9.3\% \\
Llama 4 Scout &\textbf{2.1\%} &0.0\% &0.0\% &21.0\% &4.7\% &3.5\% \\

  \bottomrule
\end{tabular}
    \caption*{\textbf{Note:} Bold values indicate the highest F1 score in each column.}

\end{table}

\begin{table}[h]
    \caption{F1 Scores for I21 Primary and All Diagnoses (Levels 3-5) by LLM}
    \label{table_I21}
    \begin{tabular}{lllllll}
    \toprule
        Model  & \makecell{Level 3\\Primary} & \makecell{Level 4\\Primary} &  \makecell{Level 5\\Primary} & \makecell{Level 3\\All} &  \makecell{Level 4\\All} & \makecell{Level 5\\All}  \\
    \midrule

Deepseek Reasoner &4.0\% &0.0\% &0.0\% &33.9\% &20.8\% &17.9\% \\
Gemini 2.0 Flash\\Thinking &1.3\% &1.1\% &0.0\% &15.2\% &8.9\% &4.6\% \\
Gemini 2.5 Pro &9.4\% &1.3\% &1.3\% &\textbf{40.5\%} &\textbf{28.0\%} &\textbf{25.7\%} \\
GPT o3 Mini &9.0\% &7.7\% &0.0\% &25.9\% &15.4\% &11.9\% \\
Qwen QWQ &14.9\% &\textbf{8.8\%} &2.4\% &29.2\% &15.1\% &5.8\% \\
    \midrule

Deepseek Chat &\textbf{41.0\%} &0.0\% &0.0\% &33.3\% &17.6\% &13.1\% \\
Gemini 2.0 Flash &1.4\% &0.8\% &0.8\% &19.0\% &10.0\% &7.9\% \\
GPT 4o &0.0\% &0.0\% &0.0\% &26.9\% &15.6\% &12.8\% \\
GPT 4o Mini &5.7\% &3.3\% &0.0\% &29.6\% &14.6\% &13.1\% \\
Llama 3.3 Versatile &15.6\% &7.9\% &\textbf{4.6\%} &26.1\% &14.3\% &11.7\% \\
Llama 4 Scout &0.0\% &0.0\% &0.0\% &21.4\% &7.2\% &5.5\% \\

  \bottomrule
\end{tabular}
    \caption*{\textbf{Note:} Bold values indicate the highest F1 score in each column.}

\end{table}

\begin{table}[h]
    \caption{F1 Scores for I25 Primary and All Diagnoses (Levels 3-5) by LLM}
    \label{table_I25}
    \begin{tabular}{lllllll}
    \toprule
        Model  & \makecell{Level 3\\Primary} & \makecell{Level 4\\Primary} &  \makecell{Level 5\\Primary} & \makecell{Level 3\\All} &  \makecell{Level 4\\All} & \makecell{Level 5\\All}  \\
    \midrule

Deepseek Reasoner &53.6\% &49.1\% &20.8\% &38.4\% &26.8\% &22.1\% \\
Gemini 2.0 Flash\\Thinking &40.4\% &35.5\% &2.9\% &17.9\% &12.7\% &5.9\% \\
Gemini 2.5 Pro &47.7\% &45.3\% &\textbf{23.7\%} &\textbf{46.4\%} &\textbf{35.1\%} &\textbf{30.5\%} \\
GPT o3 Mini &\textbf{56.5\%} &\textbf{54.5\%} &21.9\% &29.1\% &20.9\% &14.5\% \\
Qwen QWQ &29.2\% &17.7\% &1.2\% &28.5\% &16.9\% &5.9\% \\
\midrule
Deepseek Chat &19.0\% &13.7\% &4.3\% &33.1\% &23.2\% &16.9\% \\
Gemini 2.0 Flash &22.2\% &21.4\% &7.4\% &18.6\% &12.4\% &7.3\% \\
GPT 4o &34.9\% &33.8\% &8.6\% &30.5\% &21.3\% &15.7\% \\
GPT 4o Mini &42.5\% &39.8\% &16.3\% &30.8\% &18.6\% &14.3\% \\
Llama 3.3 Versatile &32.4\% &25.7\% &8.1\% &28.6\% &18.2\% &12.6\% \\
Llama 4 Scout &6.7\% &0.4\% &0.3\% &25.0\% &10.4\% &7.3\% \\

  \bottomrule
\end{tabular}
    \caption*{\textbf{Note:} Bold values indicate the highest F1 score in each column.}

\end{table}

\begin{table}[h]
    \caption{F1 Scores for Y92 Primary and All Diagnoses (Levels 3-5) by LLM}
    \label{table_Y92}
    \begin{tabular}{lllllll}
    \toprule
        Model  & \makecell{Level 3\\Primary} & \makecell{Level 4\\Primary} &  \makecell{Level 5\\Primary} & \makecell{Level 3\\All} &  \makecell{Level 4\\All} & \makecell{Level 5\\All}  \\
    \midrule
Deepseek Reasoner &0.0\% &0.0\% &0.0\% &25.9\% &\textbf{16.2\%} &\textbf{12.3\%} \\
Gemini 2.0 Flash\\Thinking &1.3\% &0.3\% &0.0\% &11.6\% &6.2\% &3.3\% \\
Gemini 2.5 Pro &3.5\% &0.8\% &0.8\% &\textbf{26.1\%} &15.5\% &8.7\% \\
GPT o3 Mini &0.0\% &0.0\% &0.0\% &15.4\% &9.3\% &5.5\% \\
Qwen QWQ &0.0\% &0.0\% &0.0\% &18.3\% &6.3\% &2.1\% \\
    \midrule
Deepseek Chat &0.0\% &0.0\% &0.0\% &22.0\% &12.9\% &10.3\% \\
Gemini 2.0 Flash &\textbf{5.3}\% &\textbf{4.1\%} &\textbf{2.5\%} &15.1\% &8.2\% &5.9\% \\
GPT 4o &3.3\% &1.2\% &1.2\% &5.1\% &3.3\% &3.3\% \\
GPT 4o Mini &0.0\% &0.0\% &0.0\% &7.3\% &4.3\% &1.7\% \\
Llama 3.3 Versatile &0.0\% &0.0\% &0.0\% &14.0\% &6.6\% &5.6\% \\
Llama 4 Scout &4.7\% &1.0\% &0.0\% &12.3\% &5.1\% &3.1\% \\

  \bottomrule
\end{tabular}
    \caption*{\textbf{Note:} Bold values indicate the highest F1 score in each column.}

\end{table}

\begin{table}[h]
    \caption{F1 Scores for Z51 Primary and All Diagnoses (Levels 3-5) by LLM}
    \label{table_Z51}
    \begin{tabular}{lllllll}
    \toprule
        Model  & \makecell{Level 3\\Primary} & \makecell{Level 4\\Primary} &  \makecell{Level 5\\Primary} & \makecell{Level 3\\All} &  \makecell{Level 4\\All} & \makecell{Level 5\\All}  \\
    \midrule
Deepseek Reasoner &0.0\% &0.0\% &0.0\% &30.3\% &23.6\% &16.3\% \\
Gemini 2.0 Flash\\Thinking &0.0\% &0.0\% &0.0\% &16.2\% &11.7\% &3.8\% \\
Gemini 2.5 Pro &0.0\% &0.0\% &0.0\% &\textbf{38.5\%} &\textbf{29.9\%} &\textbf{22.4\%} \\
GPT o3 Mini &0.0\% &0.0\% &0.0\% &19.1\% &13.7\% &5.0\% \\
Qwen QWQ &\textbf{0.2\%} &\textbf{0.2\%} &0.0\% &20.7\% &8.4\% &3.2\% \\
    \midrule
Deepseek Chat &0.0\% &0.0\% &0.0\% &23.6\% &15.1\% &11.7\% \\
Gemini 2.0 Flash &0.0\% &0.0\% &0.0\% &18.9\% &11.2\% &7.5\% \\
GPT 4o &0.0\% &0.0\% &0.0\% &20.3\% &14.8\% &9.8\% \\
GPT 4o Mini &0.0\% &0.0\% &0.0\% &21.1\% &13.5\% &9.5\% \\
Llama 3.3 Versatile &0.0\% &0.0\% &0.0\% &16.2\% &9.5\% &6.7\% \\
Llama 4 Scout &0.0\% &0.0\% &0.0\% &10.7\% &4.9\% &2.9\% \\

  \bottomrule
\end{tabular}
    \caption*{\textbf{Note:} Bold values indicate the highest F1 score in each column.}

\end{table}

\begin{table}[h]
    \caption{F1 Scores for Z79 Primary and All Diagnoses (Levels 3-5) by LLM}
    \label{table_Z79}
    \begin{tabular}{lllllll}
    \toprule
        Model  & \makecell{Level 3\\Primary} & \makecell{Level 4\\Primary} &  \makecell{Level 5\\Primary} & \makecell{Level 3\\All} &  \makecell{Level 4\\All} & \makecell{Level 5\\All}  \\
    \midrule

Deepseek Reasoner &\textbf{19.6\%} &11.5\% &\textbf{7.6\%} &28.9\% &16.2\% &13.4\% \\
Gemini 2.0 Flash\\Thinking &10.8\% &6.9\% &1.1\% &11.5\% &6.8\% &3.2\% \\
Gemini 2.5 Pro &16.1\% &\textbf{12.0\%} &6.7\% &\textbf{37.5\%} &\textbf{24.3\%} &\textbf{20.8\%} \\
GPT o3 Mini &0.0\% &0.0\% &0.0\% &16.5\% &8.5\% &5.9\% \\
Qwen QWQ &9.4\% &3.0\% &0.1\% &19.8\% &7.2\% &3.2\% \\
    \midrule
Deepseek Chat &11\% &4.5\% &2.5\% &27.4\% &13.2\% &9.3\% \\
Gemini 2.0 Flash &4.8\% &4.4\% &1.4\% &15.3\% &8.1\% &6.5\% \\
GPT 4o &9.3\% &8.3\% &3.3\% &20.3\% &10.3\% &8.5\% \\
GPT 4o Mini &10.8\% &6.5\% &0.0\% &24.3\% &12.1\% &10.1\% \\
Llama 3.3 Versatile &0.0\% &0.0\% &0.0\% &18.1\% &6.3\% &5.5\% \\
Llama 4 Scout &4.7\% &1.0\% &0.3\% &14.2\% &4.9\% &3.4\% \\

  \bottomrule
\end{tabular}
    \caption*{\textbf{Note:} Bold values indicate the highest F1 score in each column.}

\end{table}

\begin{table}[h]
    \caption{F1 Scores for Z87 Primary and All Diagnoses (Levels 3-5) by LLM}
    \label{table_Z87}
    \begin{tabular}{lllllll}
    \toprule
        Model  & \makecell{Level 3\\Primary} & \makecell{Level 4\\Primary} &  \makecell{Level 5\\Primary} & \makecell{Level 3\\All} &  \makecell{Level 4\\All} & \makecell{Level 5\\All}  \\
    \midrule
Deepseek Reasoner &3.8\% &2.0\% &0.0\% &32.6\% &23.9\% &19.8\% \\
Gemini 2.0 Flash\\Thinking &14.7\% &12.3\% &1.6\% &12.8\% &6.9\% &3.6\% \\
Gemini 2.5 Pro &\textbf{25.9\%} &\textbf{23.1\%} &\textbf{11.4\%} &\textbf{35.3\%} &\textbf{27.2\%} &\textbf{24.5\%} \\
GPT o3 Mini &0.0\% &0.0\% &0.0\% &22.0\% &11.1\% &6.8\% \\
Qwen QWQ &6.0\% &3.3\% &0.2\% &20.5\% &8.6\% &3.6\% \\
    \midrule
Deepseek Chat &0.0\% &0.0\% &0.0\% &26.1\% &15.3\% &10.7\% \\
Gemini 2.0 Flash &11.8\% &9.7\% &3.2\% &17.7\% &9.8\% &6.9\% \\
GPT 4o &15.6\% &10.7\% &3.1\% &21.5\% &11.0\% &8.7\% \\
GPT 4o Mini &0.0\% &0.0\% &0.0\% &23.2\% &14.7\% &10.8\% \\
Llama 3.3 Versatile &0.0\% &0.0\% &0.0\% &19.7\% &7.3\% &6.5\% \\
Llama 4 Scout &6.5\% &0.7\% &0.2\% &15.3\% &5.5\% &3.6\% \\

  \bottomrule
\end{tabular}
    \caption*{\textbf{Note:} Bold values indicate the highest F1 score in each column.}

\end{table}

\clearpage

\subsection{Charts}

%%% A41 %%%%%%%%%%%%%%%%%%%%%%%%%%%%%%%%%%%%%%%%%%%%%%%%%%%
\begin{figure}[h]
\centering
\begin{subfigure}{.6\textwidth}

\includegraphics[width=1\textwidth]{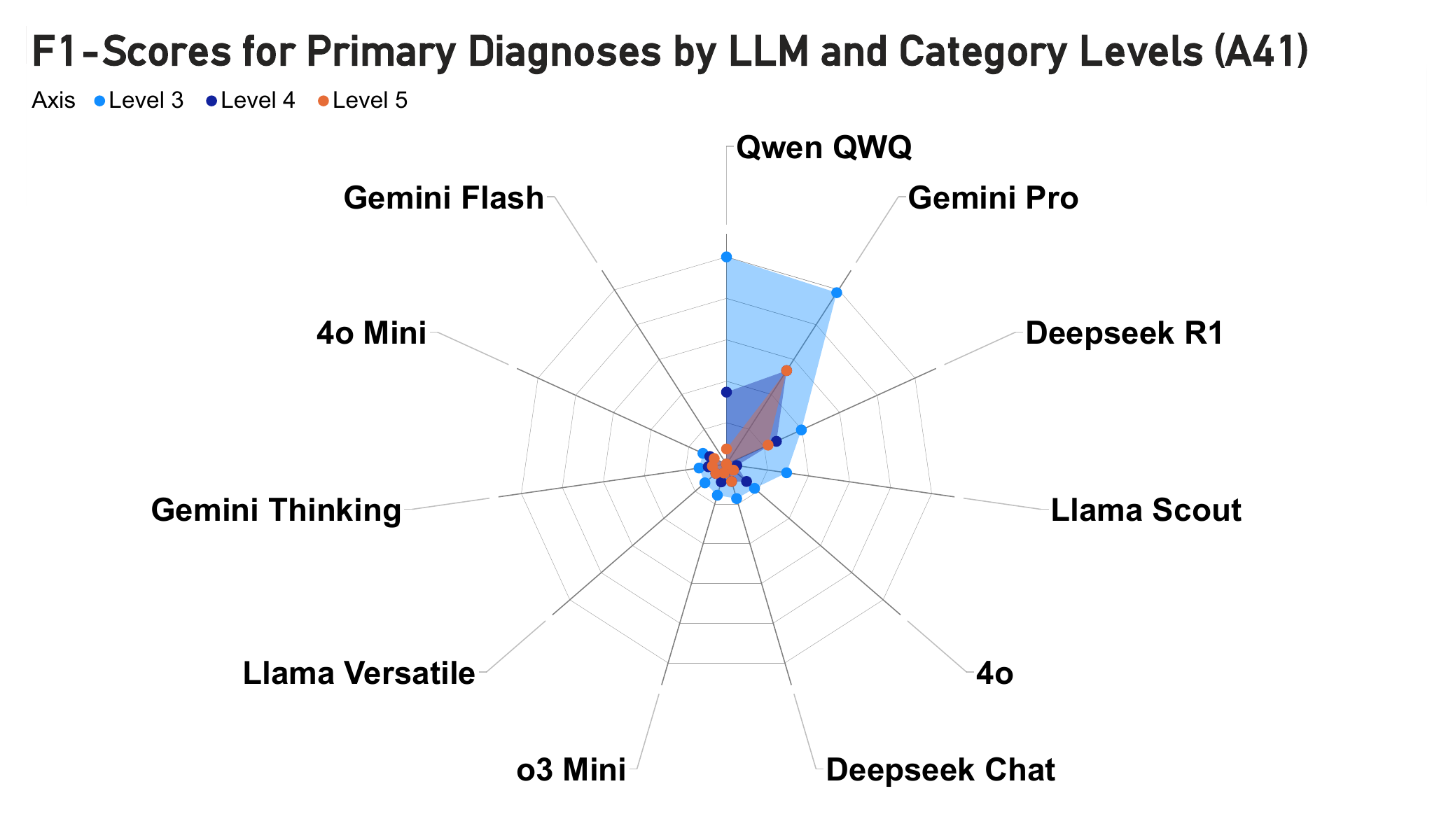}
\noindent
\caption{F1 scores for ICD-10 code A41 in primary diagnoses,\\by LLM, across ICD-10 levels 3 to 5}
\label{MiPriA41}
\end{subfigure}%
\begin{subfigure}{.6\textwidth}

\centering
\includegraphics[width=1\textwidth]{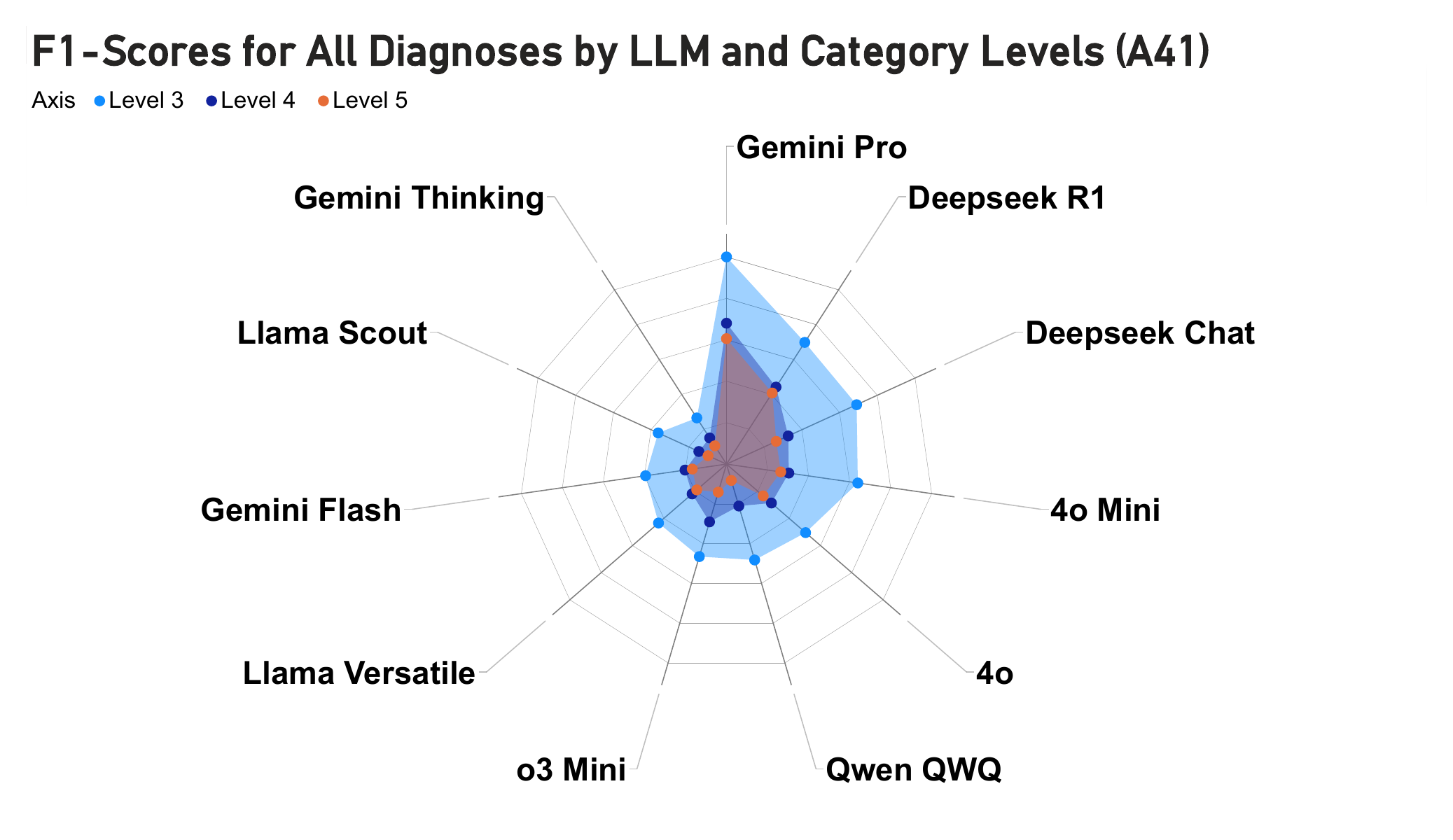}
\noindent
\caption{F1 scores for ICD-10 code A41 in all diagnoses,\\by LLM, across ICD-10 levels 3 to 5}
\label{MiAllA41}
\end{subfigure}%
\caption{F1 scores for ICD-10 code A41, by LLM, across ICD-10 levels 3 to 5}

\end{figure}

%%% E78 %%%%%%%%%%%%%%%%%%%%%%%%%%%%%%%%%%%%%%%%%%%%%%%%%%%
\begin{figure}[h]
\centering
\begin{subfigure}{.6\textwidth}

\includegraphics[width=1\textwidth]{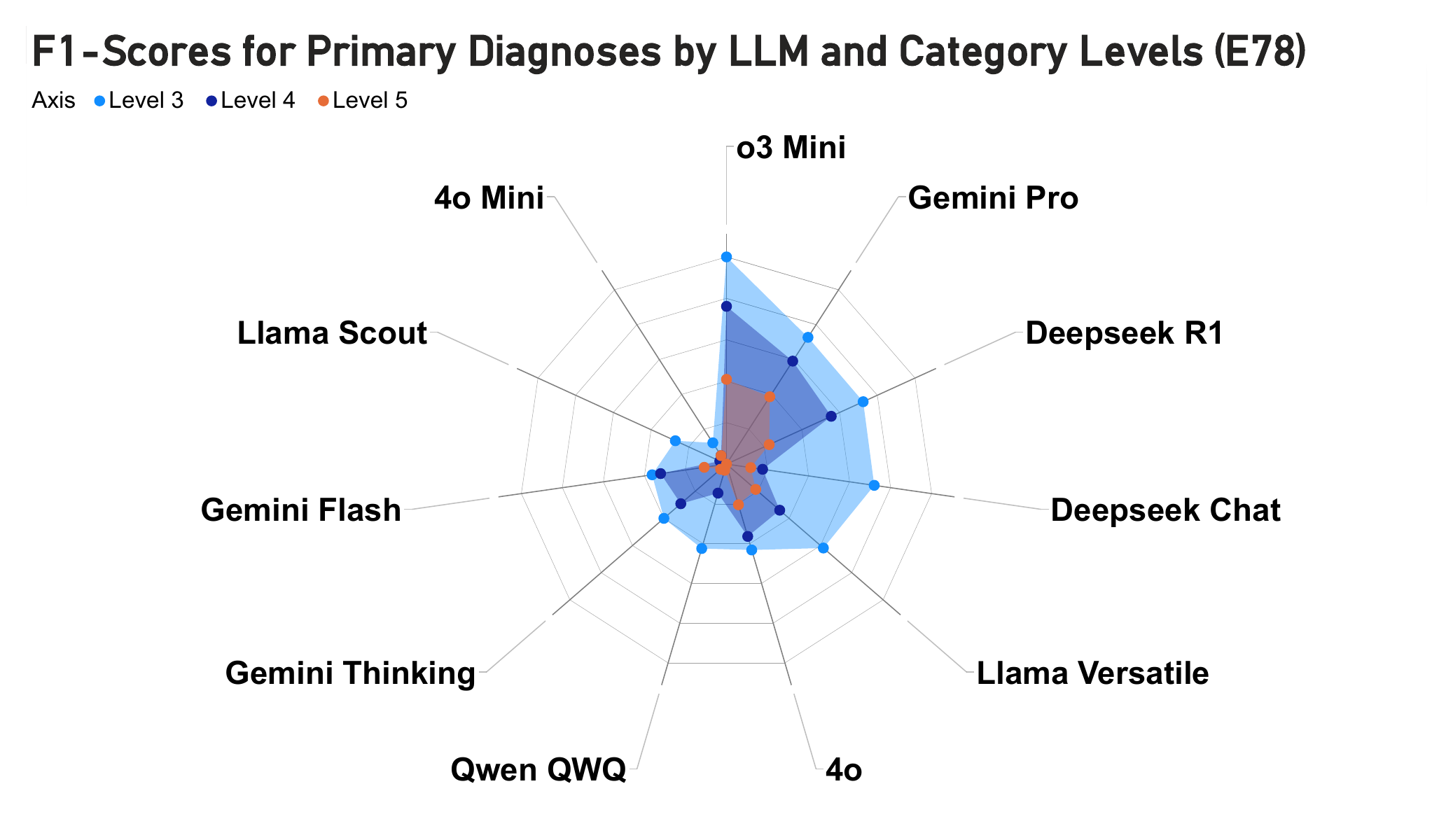}
\noindent
\caption{F1 scores for ICD-10 code E78 in primary diagnoses,\\by LLM, across ICD-10 levels 3 to 5}
\label{MiPriE78}
\end{subfigure}%
\begin{subfigure}{.6\textwidth}

\centering
\includegraphics[width=1\textwidth]{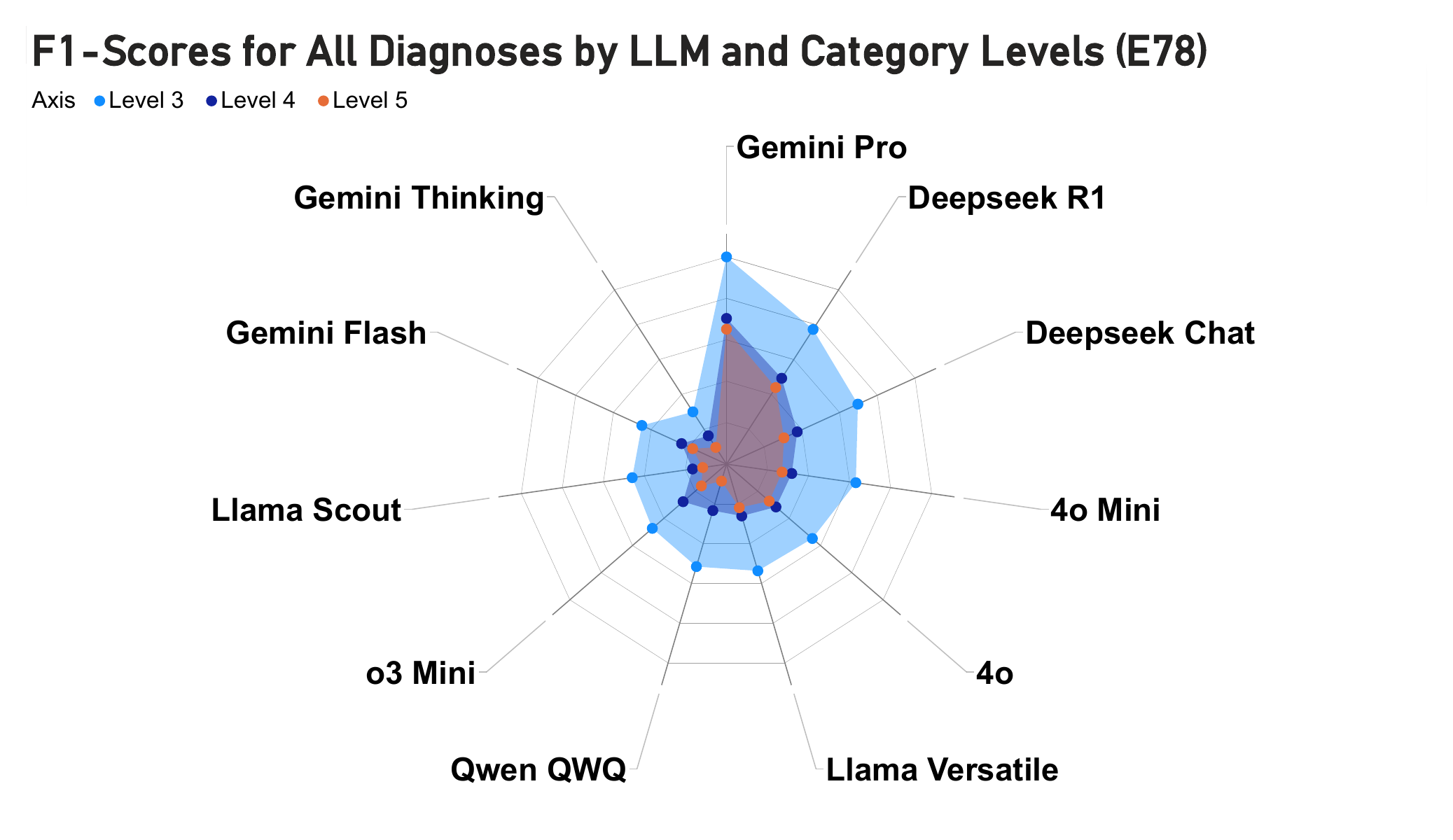}
\noindent
\caption{F1 scores for ICD-10 code E78 in all diagnoses,\\by LLM, across ICD-10 levels 3 to 5}
\label{MiAllE78}
\end{subfigure}%
\caption{F1 scores for ICD-10 code E78, by LLM, across ICD-10 levels 3 to 5}

\end{figure}

%%% I10 %%%%%%%%%%%%%%%%%%%%%%%%%%%%%%%%%%%%%%%%%%%%%%%%%%%
\begin{figure}[h]
\centering
\begin{subfigure}{.6\textwidth}

\includegraphics[width=1\textwidth]{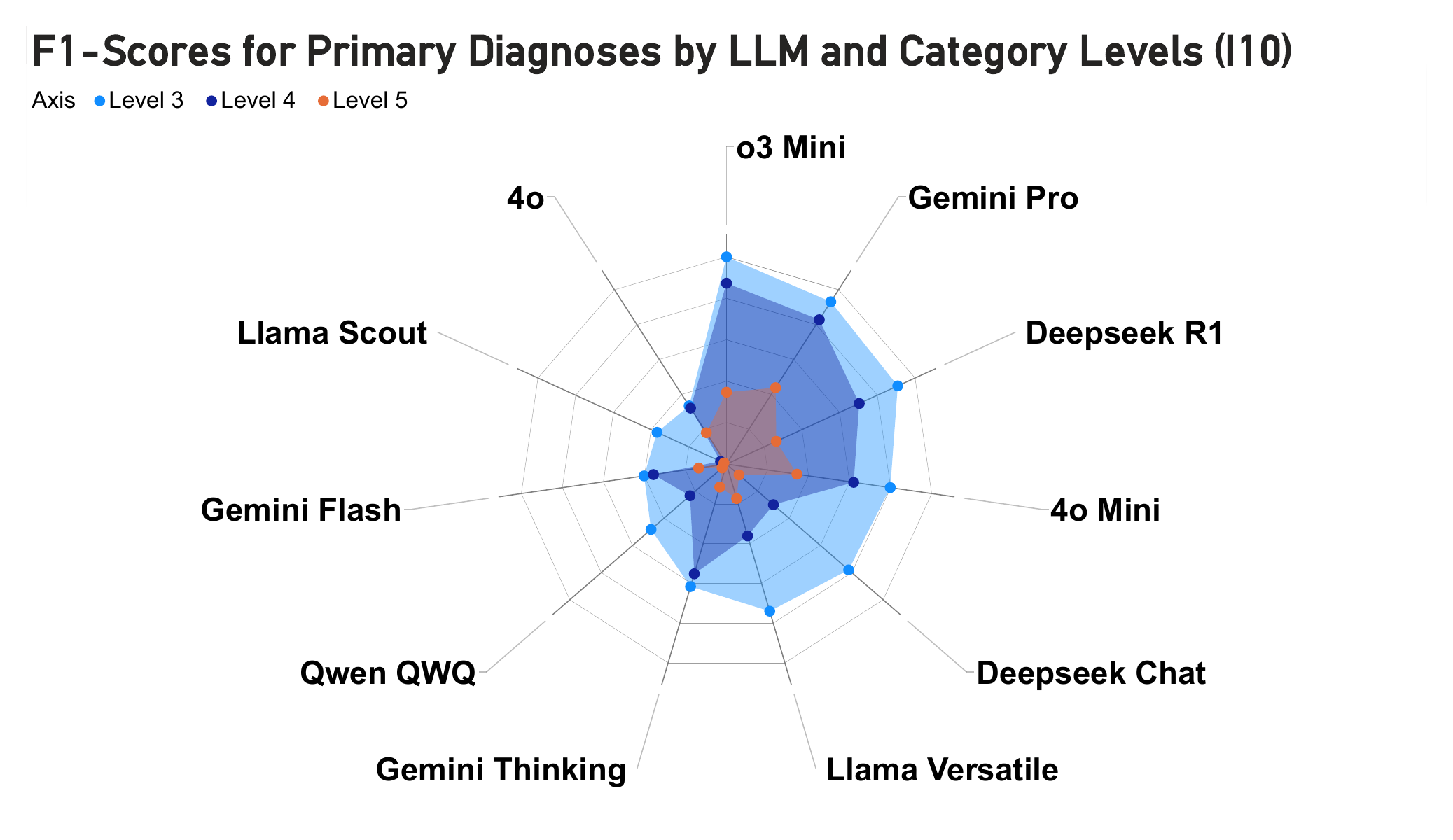}
\noindent
\caption{F1 scores for ICD-10 code I10 in primary diagnoses,\\by LLM, across ICD-10 levels 3 to 5}
\label{MiPriI10}
\end{subfigure}%
\begin{subfigure}{.6\textwidth}

\centering
\includegraphics[width=1\textwidth]{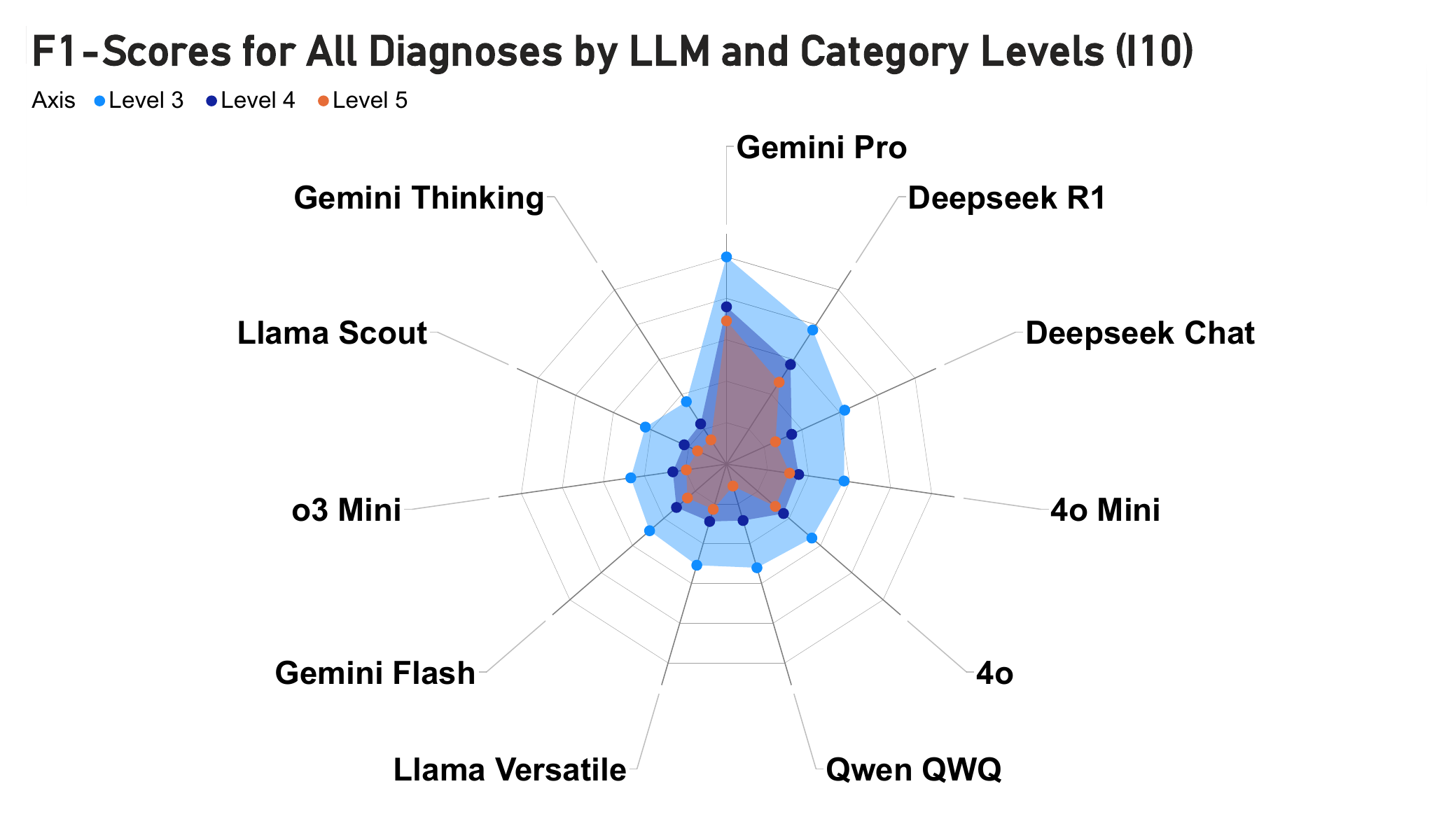}
\noindent
\caption{F1 scores for ICD-10 code I10 in all diagnoses,\\by LLM, across ICD-10 levels 3 to 5}
\label{MiAllI10}
\end{subfigure}%
\caption{F1 scores for ICD-10 code I10, by LLM, across ICD-10 levels 3 to 5}

\end{figure}

%%% I13 %%%%%%%%%%%%%%%%%%%%%%%%%%%%%%%%%%%%%%%%%%%%%%%%%%%
\begin{figure}[h]
\centering
\begin{subfigure}{.6\textwidth}

\includegraphics[width=1\textwidth]{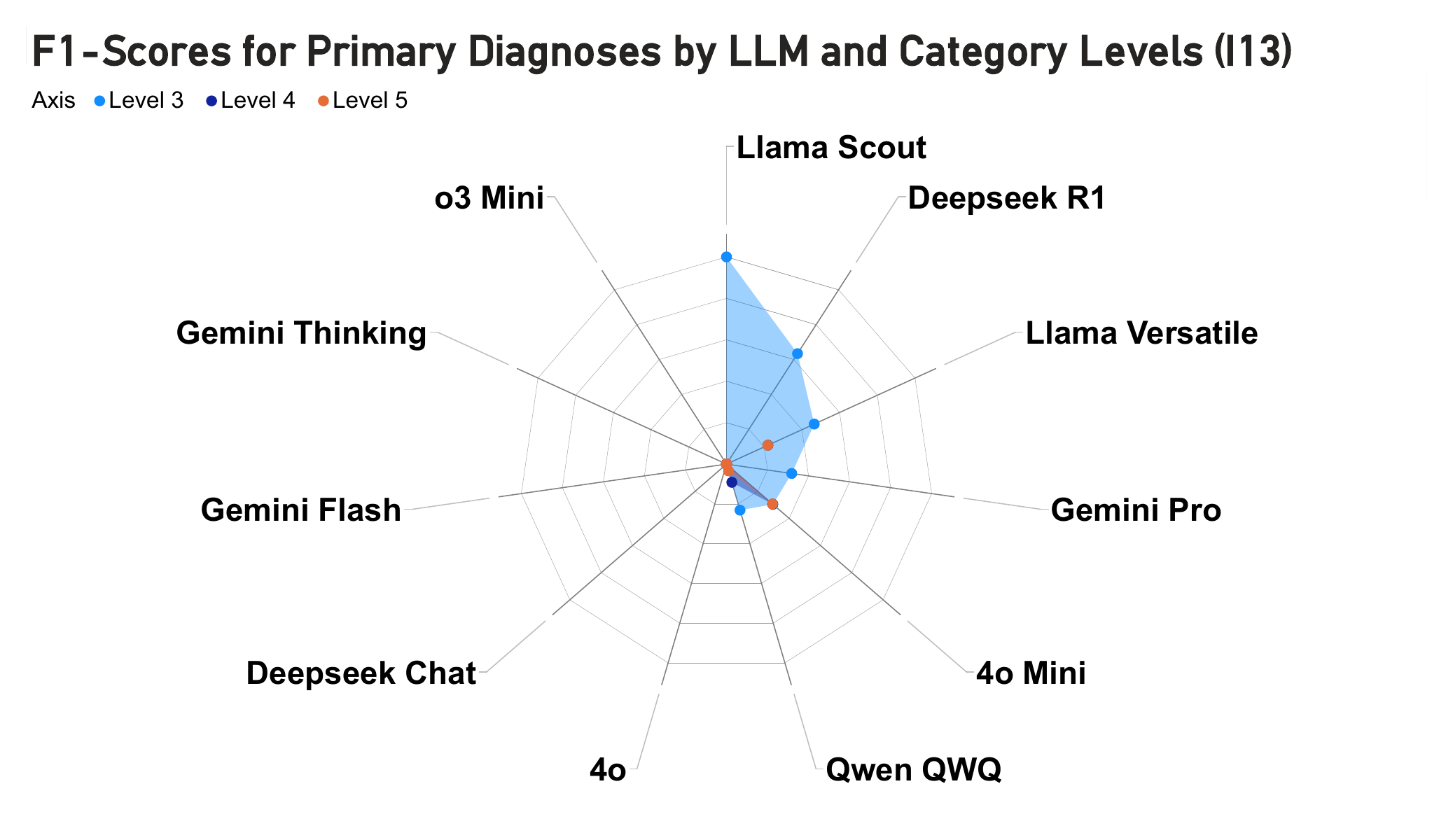}
\noindent
\caption{F1 scores for ICD-10 code I13 in primary diagnoses,\\by LLM, across ICD-10 levels 3 to 5}
\label{MiPriI13}
\end{subfigure}%
\begin{subfigure}{.6\textwidth}

\centering
\includegraphics[width=1\textwidth]{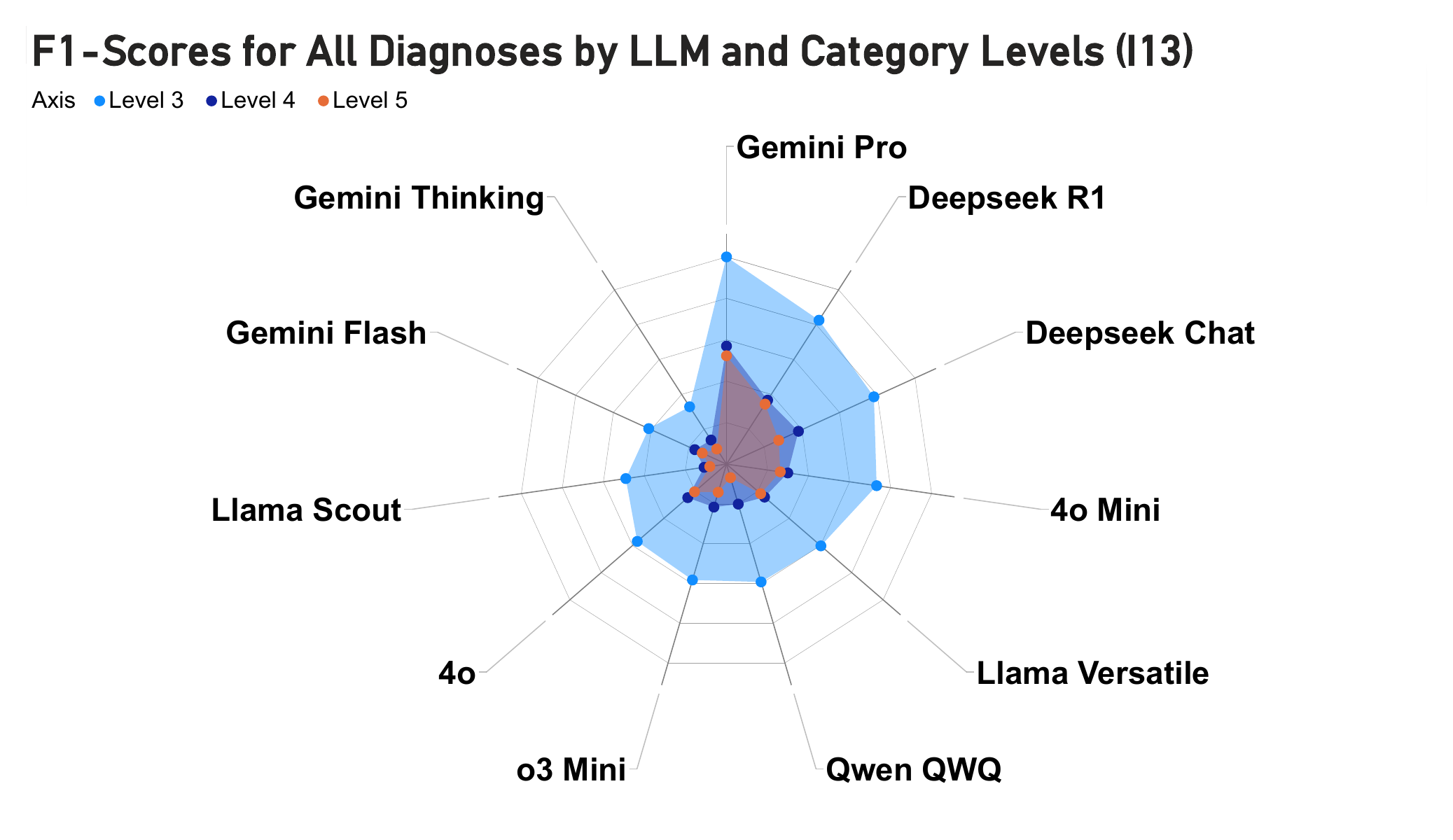}
\noindent
\caption{F1 scores for ICD-10 code I13 in all diagnoses,\\by LLM, across ICD-10 levels 3 to 5}
\label{MiAllI13}
\end{subfigure}%
\caption{F1 scores for ICD-10 code I13, by LLM, across ICD-10 levels 3 to 5}

\end{figure}

%%% I21 %%%%%%%%%%%%%%%%%%%%%%%%%%%%%%%%%%%%%%%%%%%%%%%%%%%
\begin{figure}[h]
\centering
\begin{subfigure}{.6\textwidth}

\includegraphics[width=1\textwidth]{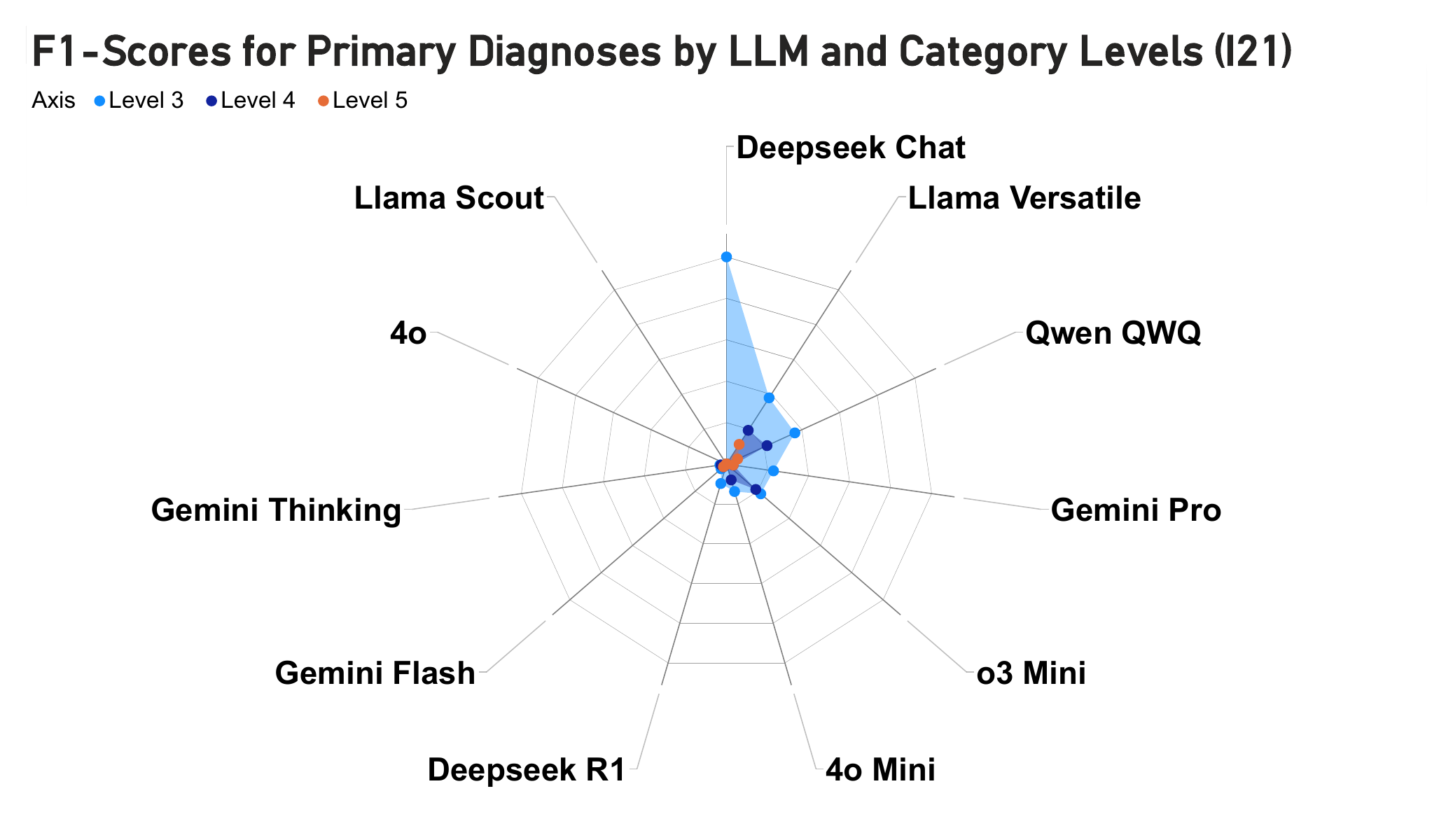}
\noindent
\caption{F1 scores for ICD-10 code I21 in primary diagnoses,\\by LLM, across ICD-10 levels 3 to 5}
\label{MiPriI21}
\end{subfigure}%
\begin{subfigure}{.6\textwidth}

\centering
\includegraphics[width=1\textwidth]{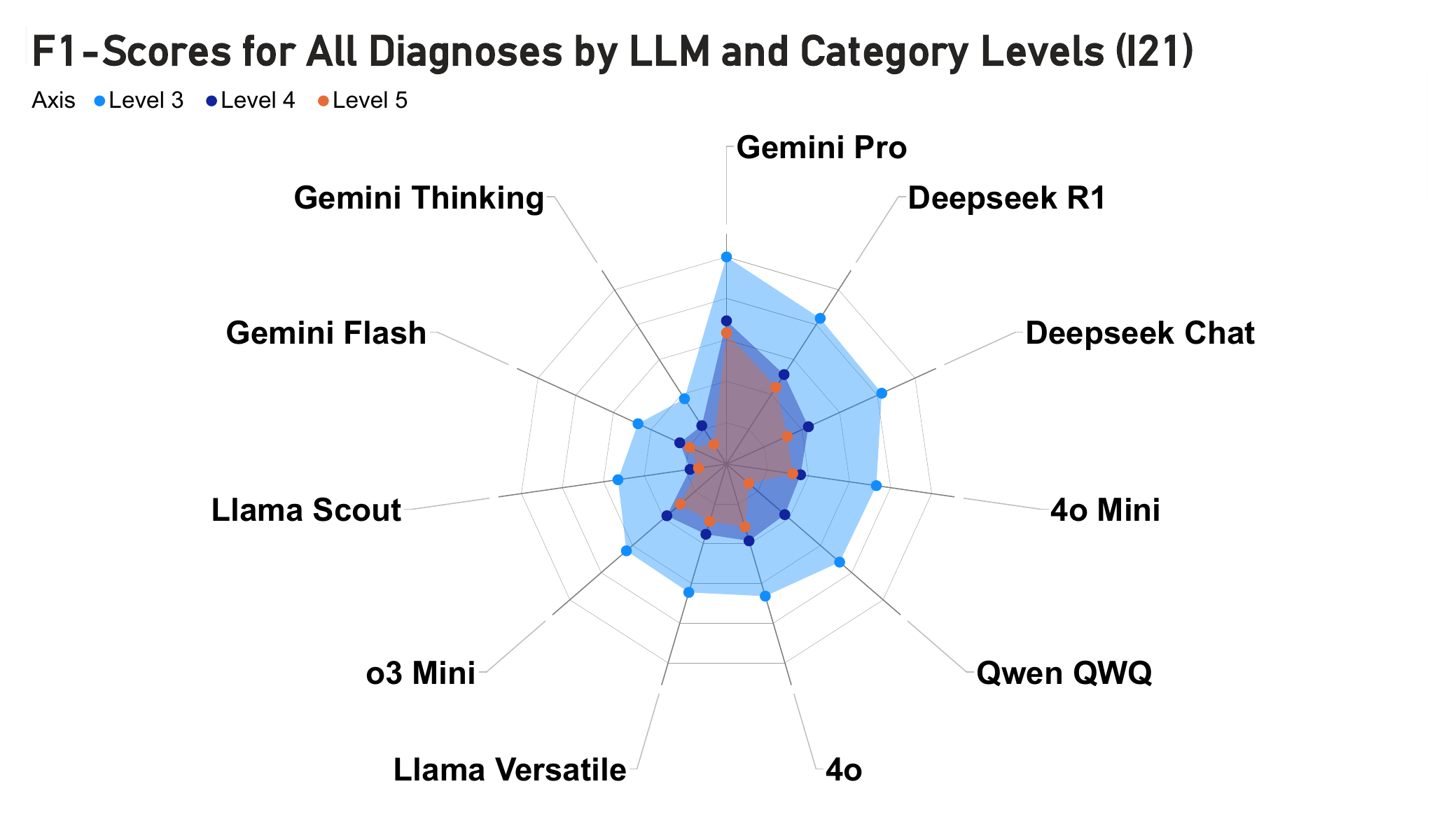}
\noindent
\caption{F1 scores for ICD-10 code I21 in all diagnoses,\\by LLM, across ICD-10 levels 3 to 5}
\label{MiAllI21}
\end{subfigure}%
\caption{F1 scores for ICD-10 code I21, by LLM, across ICD-10 levels 3 to 5}

\end{figure}

%%% I25 %%%%%%%%%%%%%%%%%%%%%%%%%%%%%%%%%%%%%%%%%%%%%%%%%%%
\begin{figure}[h]
\centering
\begin{subfigure}{.6\textwidth}

\includegraphics[width=1\textwidth]{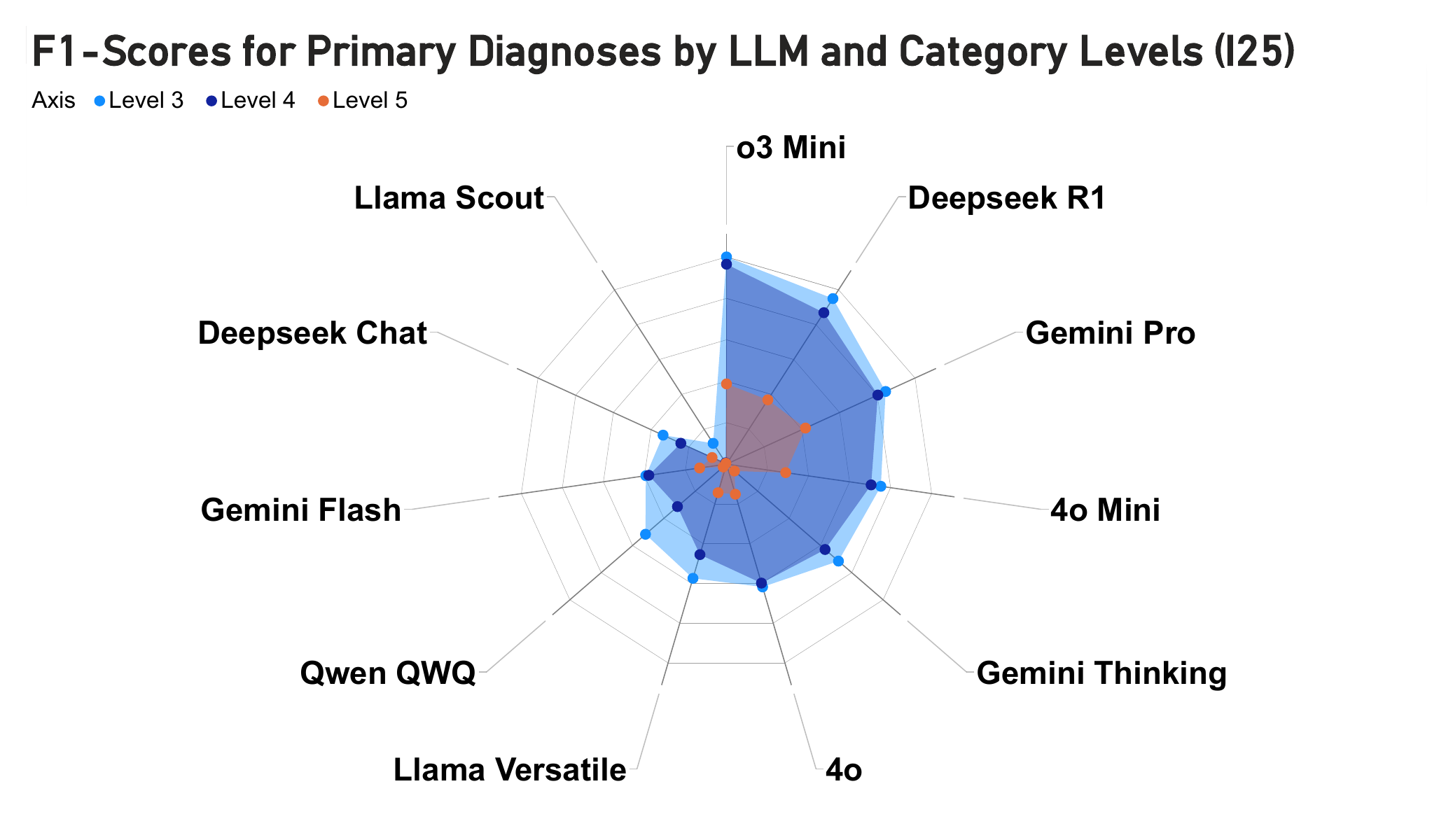}
\noindent
\caption{F1 scores for ICD-10 code I25 in primary diagnoses,\\by LLM, across ICD-10 levels 3 to 5}
\label{MiPriI25}
\end{subfigure}%
\begin{subfigure}{.6\textwidth}

\centering
\includegraphics[width=1\textwidth]{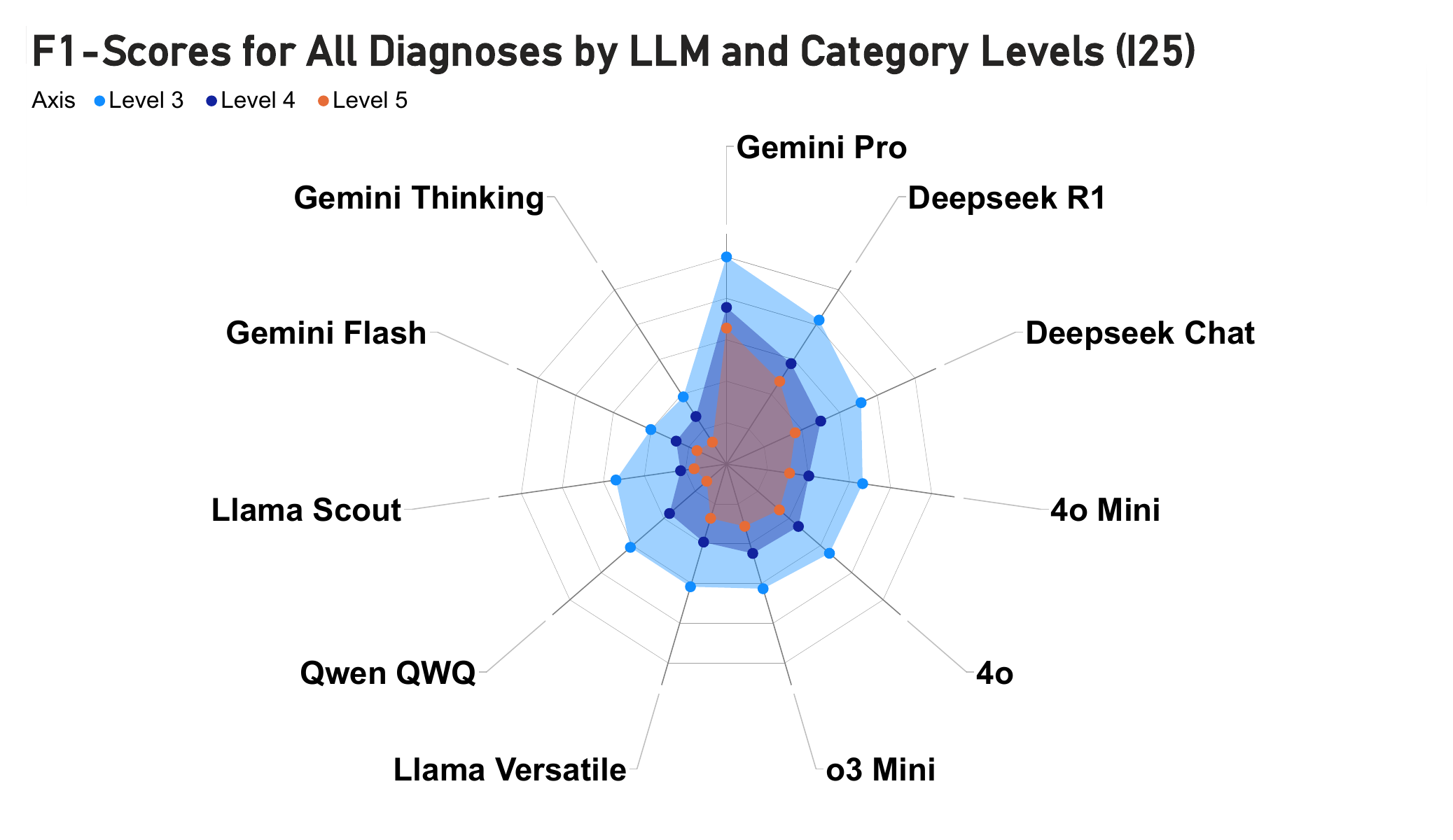}
\noindent
\caption{F1 scores for ICD-10 code I25 in all diagnoses,\\by LLM, across ICD-10 levels 3 to 5}
\label{MiAllI25}
\end{subfigure}%
\caption{F1 scores for ICD-10 code I25, by LLM, across ICD-10 levels 3 to 5}

\end{figure}

%%% Y92 %%%%%%%%%%%%%%%%%%%%%%%%%%%%%%%%%%%%%%%%%%%%%%%%%%%
\begin{figure}[h]
\centering
\begin{subfigure}{.6\textwidth}

\includegraphics[width=1\textwidth]{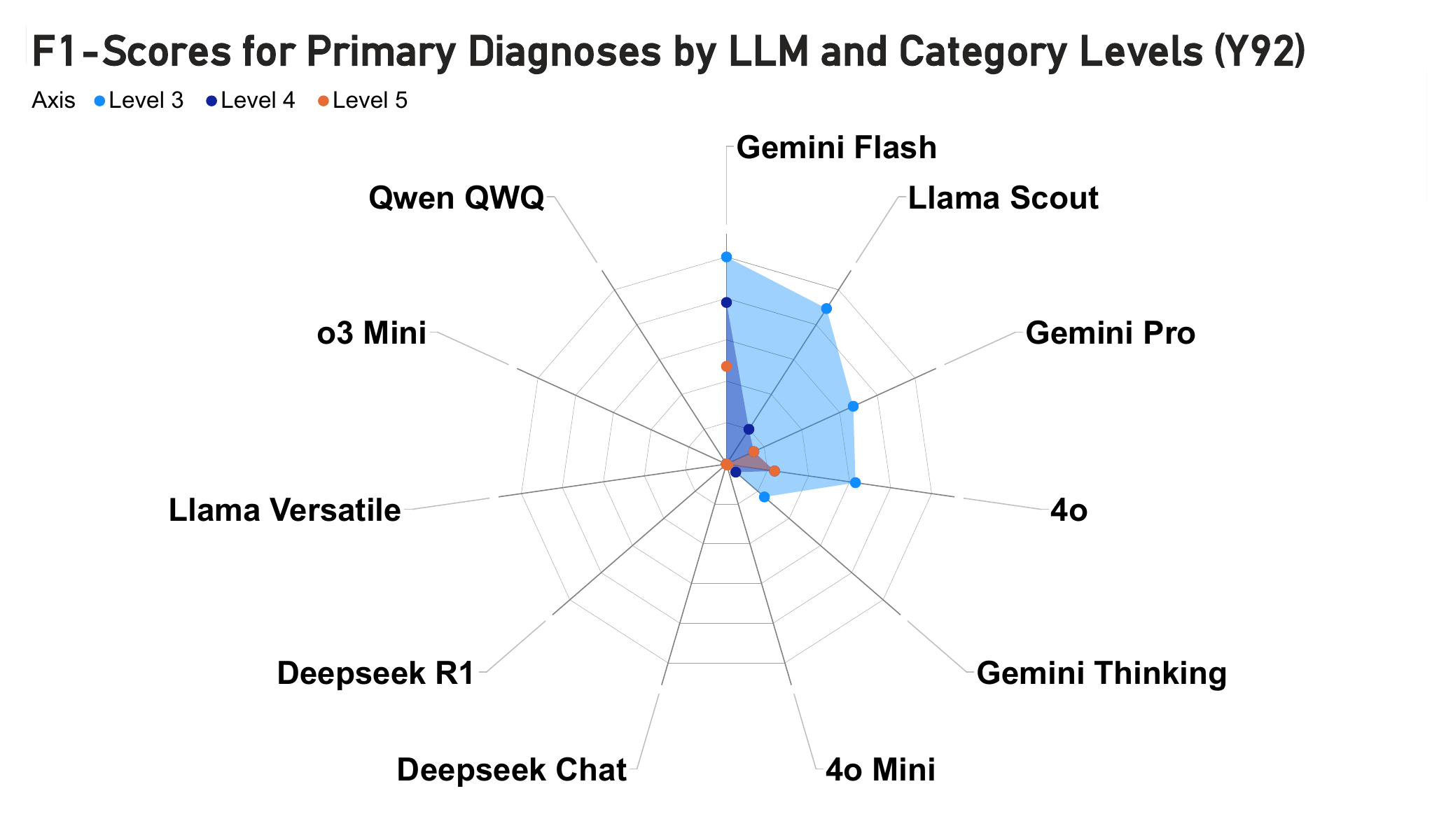}
\noindent
\caption{F1 scores for ICD-10 code Y92 in primary diagnoses,\\by LLM, across ICD-10 levels 3 to 5}
\label{MiPriY92}
\end{subfigure}%
\begin{subfigure}{.6\textwidth}

\centering
\includegraphics[width=1\textwidth]{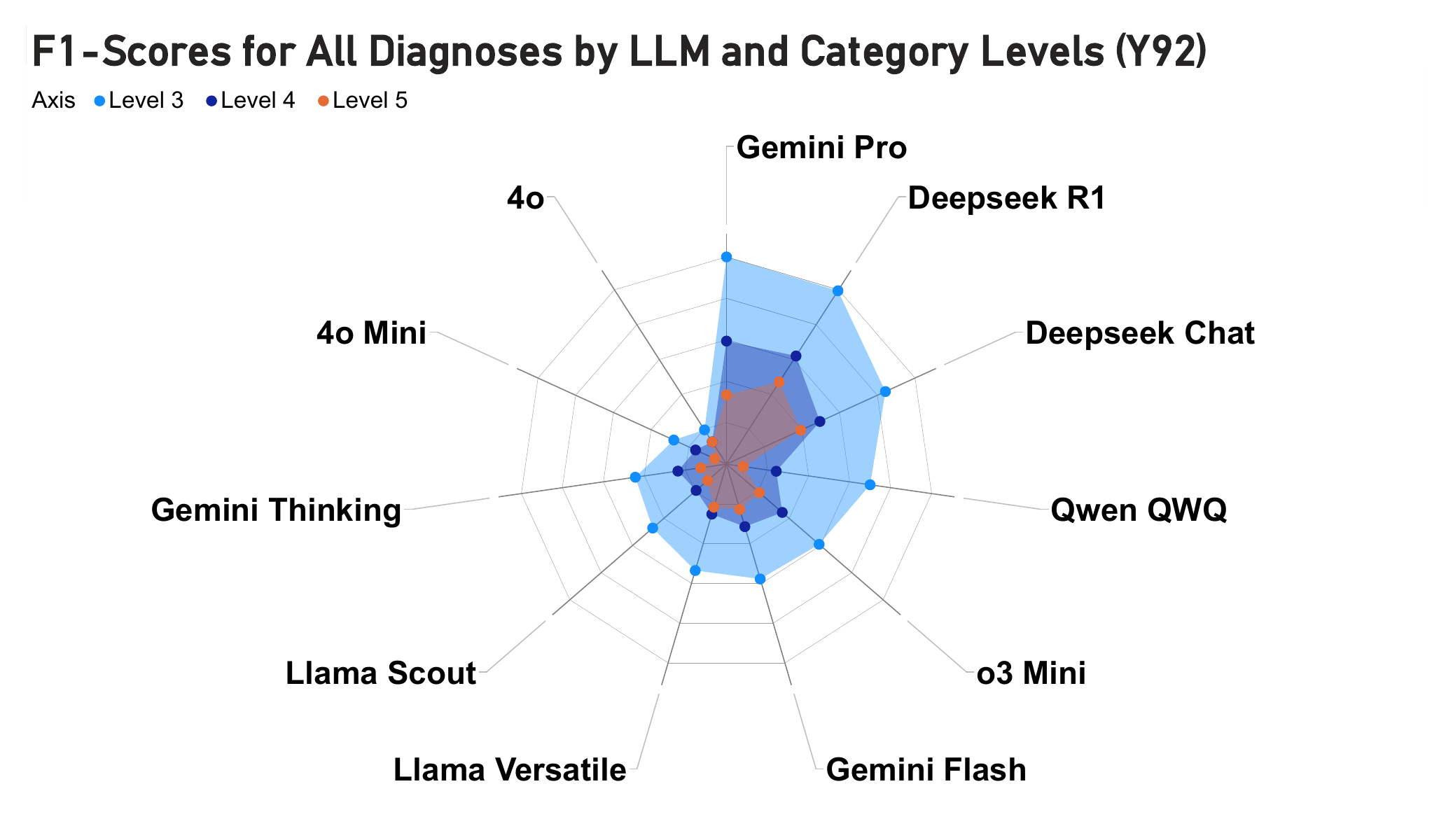}
\noindent
\caption{F1 scores for ICD-10 code Y92 in all diagnoses,\\by LLM, across ICD-10 levels 3 to 5}
\label{MiAllY92}
\end{subfigure}%
\caption{F1 scores for ICD-10 code Y92, by LLM, across ICD-10 levels 3 to 5}

\end{figure}

%%% Z51 %%%%%%%%%%%%%%%%%%%%%%%%%%%%%%%%%%%%%%%%%%%%%%%%%%%
\begin{figure}[h]
\centering
\begin{subfigure}{.6\textwidth}

\includegraphics[width=1\textwidth]{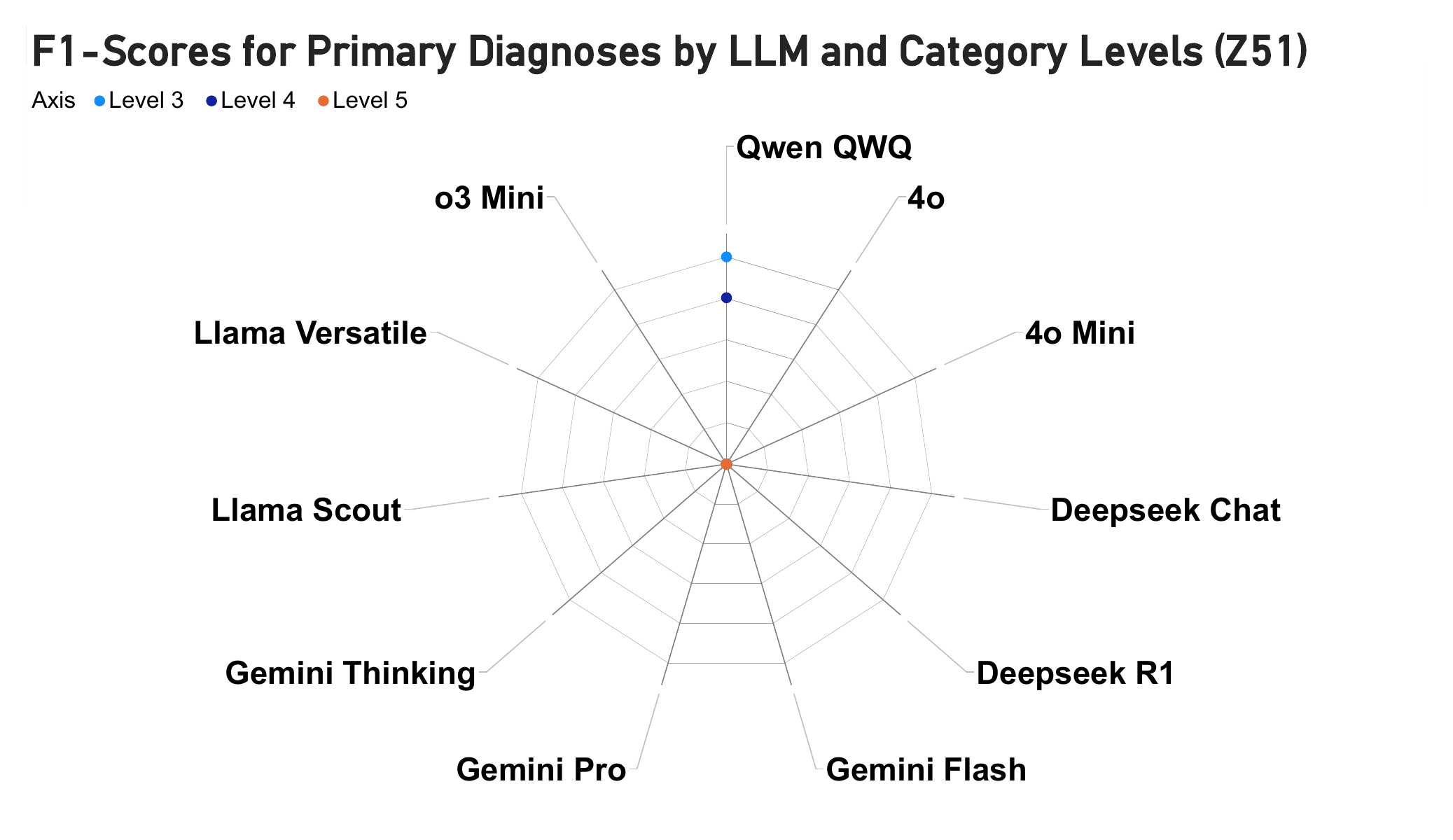}
\noindent
\caption{F1 scores for ICD-10 code Z51 in primary diagnoses,\\by LLM, across ICD-10 levels 3 to 5}
\label{MiPriZ51}
\end{subfigure}%
\begin{subfigure}{.6\textwidth}

\centering
\includegraphics[width=1\textwidth]{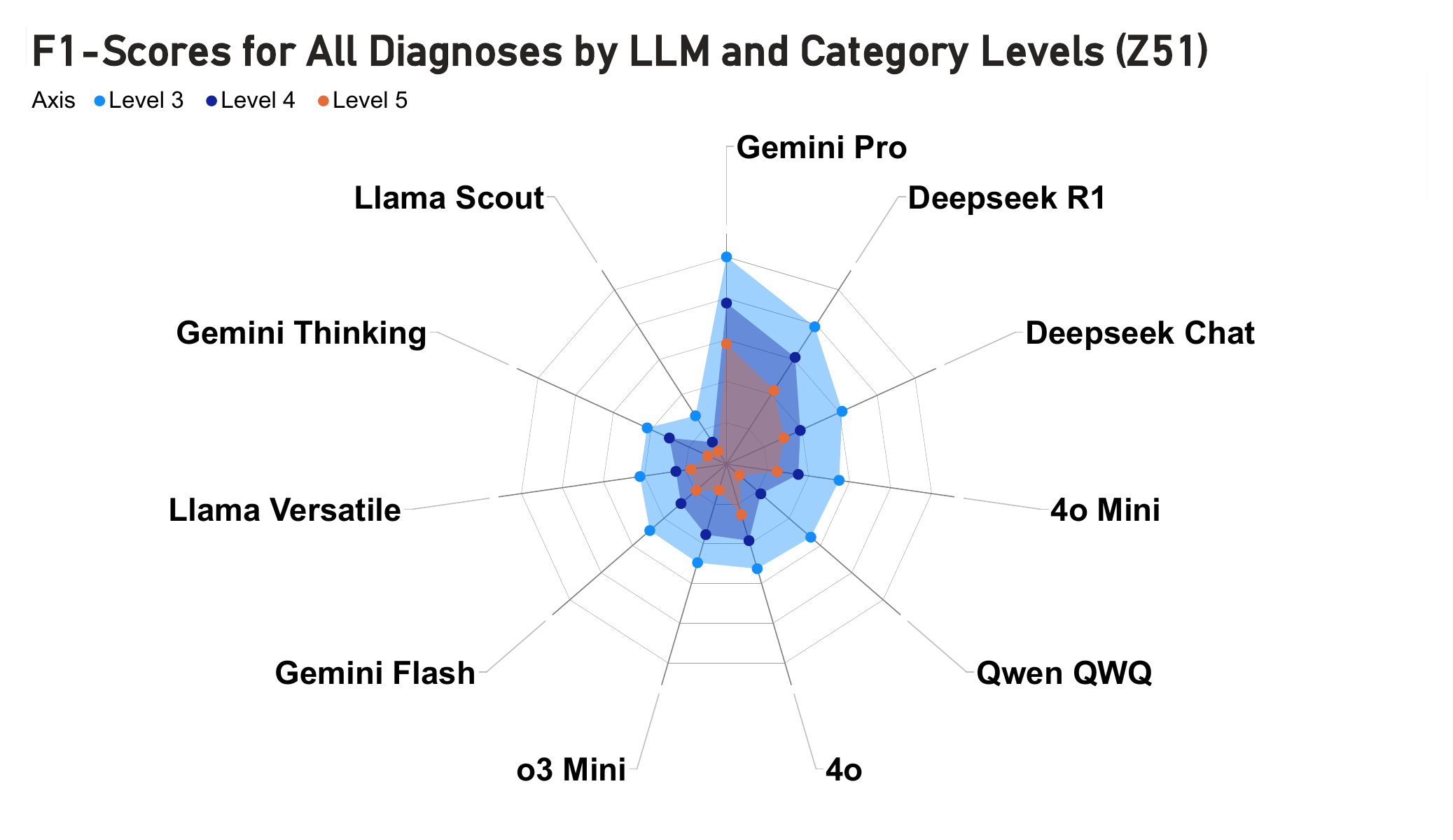}
\noindent
\caption{F1 scores for ICD-10 code Z51 in all diagnoses,\\by LLM, across ICD-10 levels 3 to 5}
\label{MiAllZ51}
\end{subfigure}%
\caption{F1 scores for ICD-10 code Z51, by LLM, across ICD-10 levels 3 to 5}

\end{figure}

%%% Z79 %%%%%%%%%%%%%%%%%%%%%%%%%%%%%%%%%%%%%%%%%%%%%%%%%%%
\begin{figure}[h]
\centering
\begin{subfigure}{.6\textwidth}

\includegraphics[width=1\textwidth]{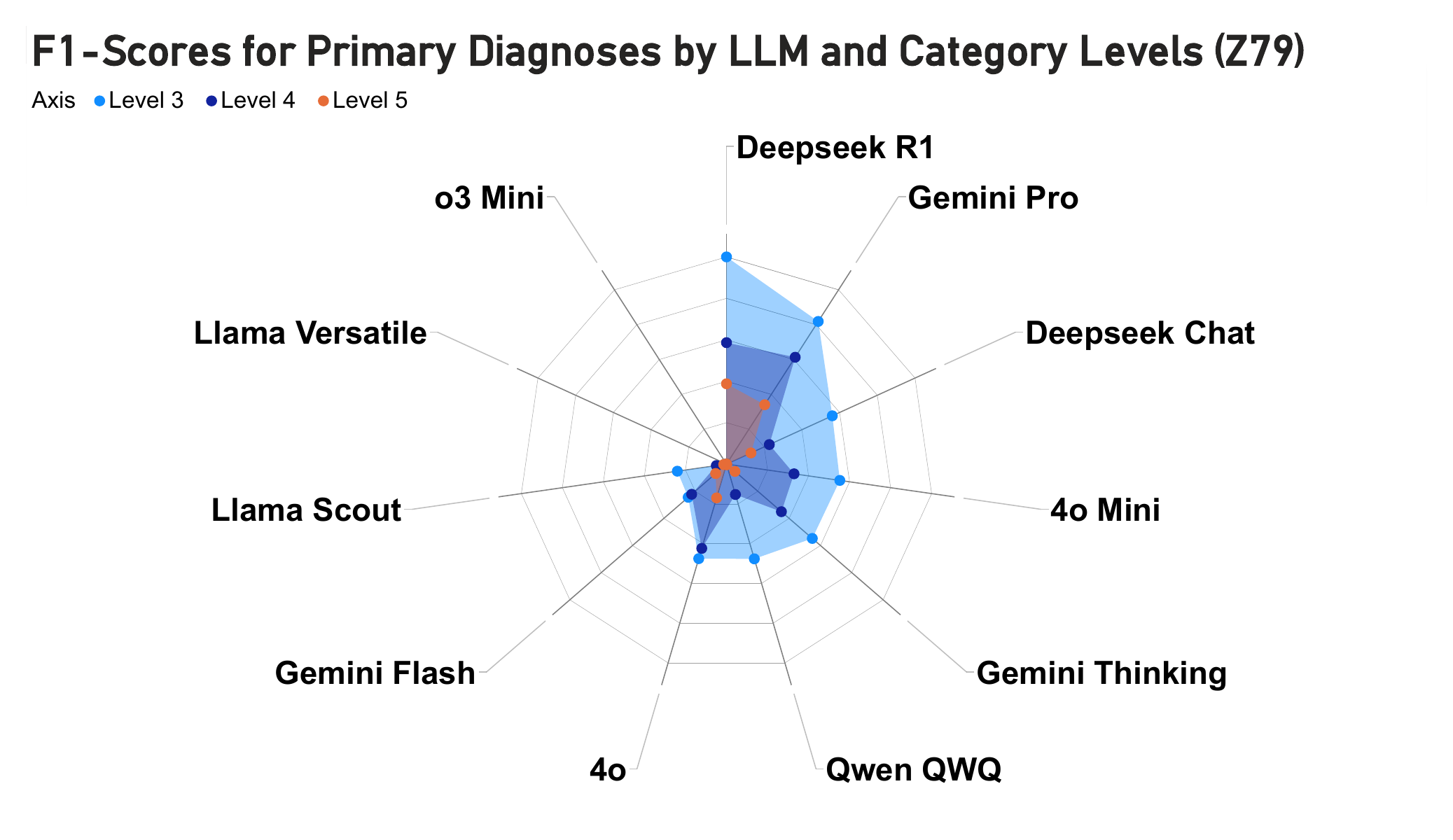}
\noindent
\caption{F1 scores for ICD-10 code Z79 in primary diagnoses,\\by LLM, across ICD-10 levels 3 to 5}
\label{MiPriZ79}
\end{subfigure}%
\begin{subfigure}{.6\textwidth}

\centering
\includegraphics[width=1\textwidth]{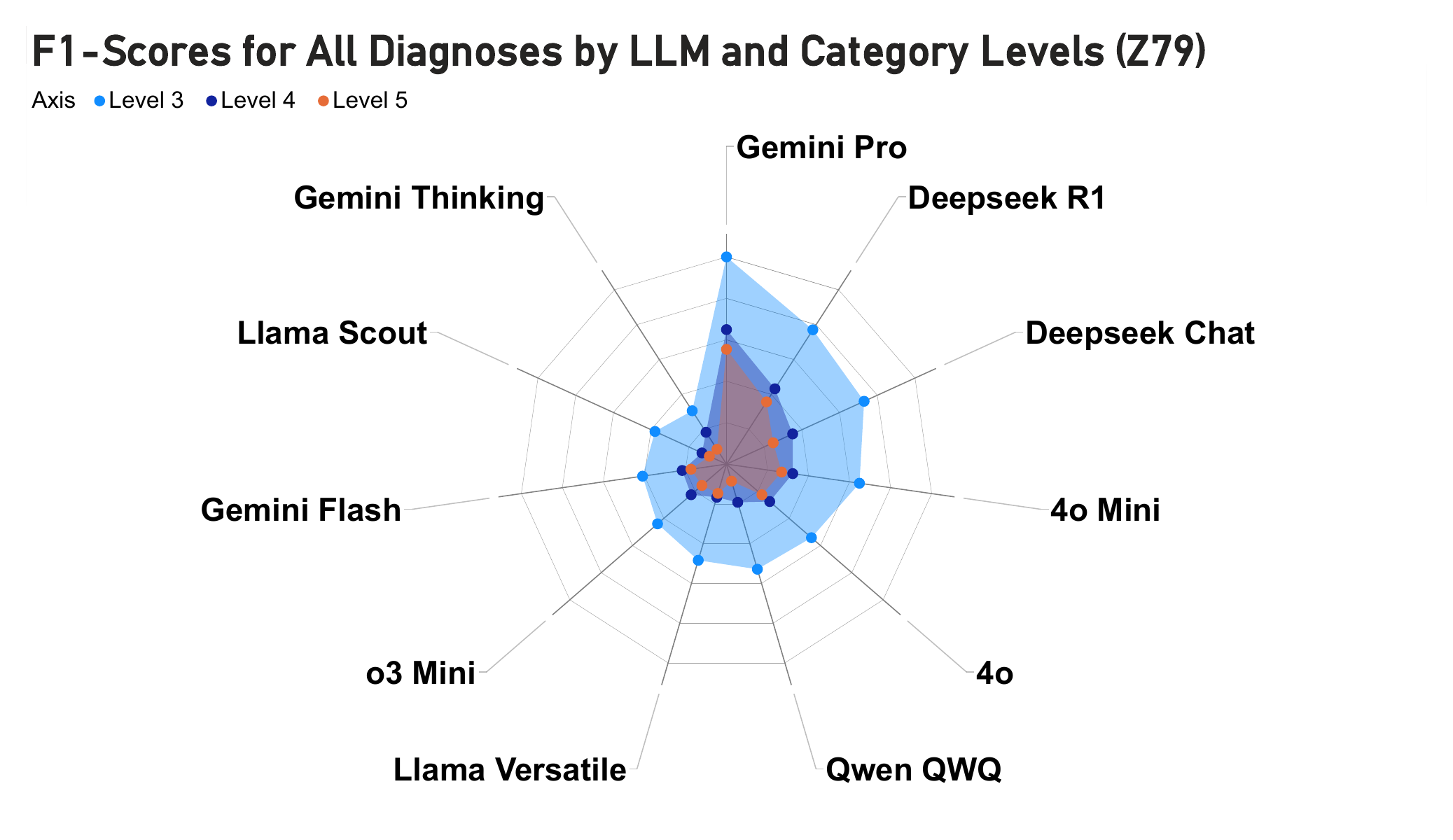}
\noindent
\caption{F1 scores for ICD-10 code Z79 in all diagnoses,\\by LLM, across ICD-10 levels 3 to 5}
\label{MiAllZ79}
\end{subfigure}%
\caption{F1 scores for ICD-10 code Z79, by LLM, across ICD-10 levels 3 to 5}

\end{figure}

%%% Z87 %%%%%%%%%%%%%%%%%%%%%%%%%%%%%%%%%%%%%%%%%%%%%%%%%%%
\begin{figure}[h]
\centering
\begin{subfigure}{.6\textwidth}

\includegraphics[width=1\textwidth]{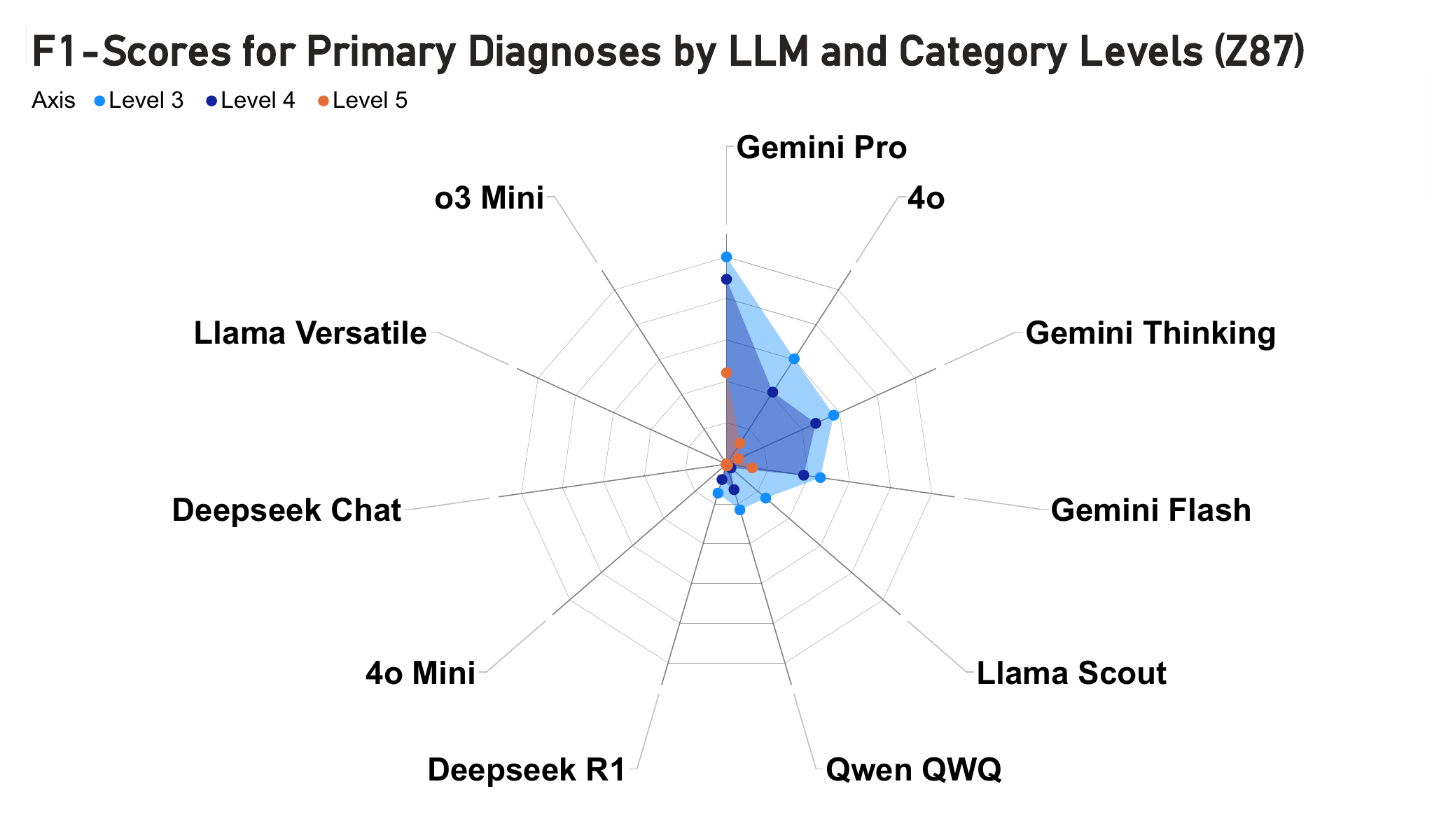}
\noindent
\caption{F1 scores for ICD-10 code Z87 in primary diagnoses,\\by LLM, across ICD-10 levels 3 to 5}
\label{MiPriZ87}
\end{subfigure}%
\begin{subfigure}{.6\textwidth}

\centering
\includegraphics[width=1\textwidth]{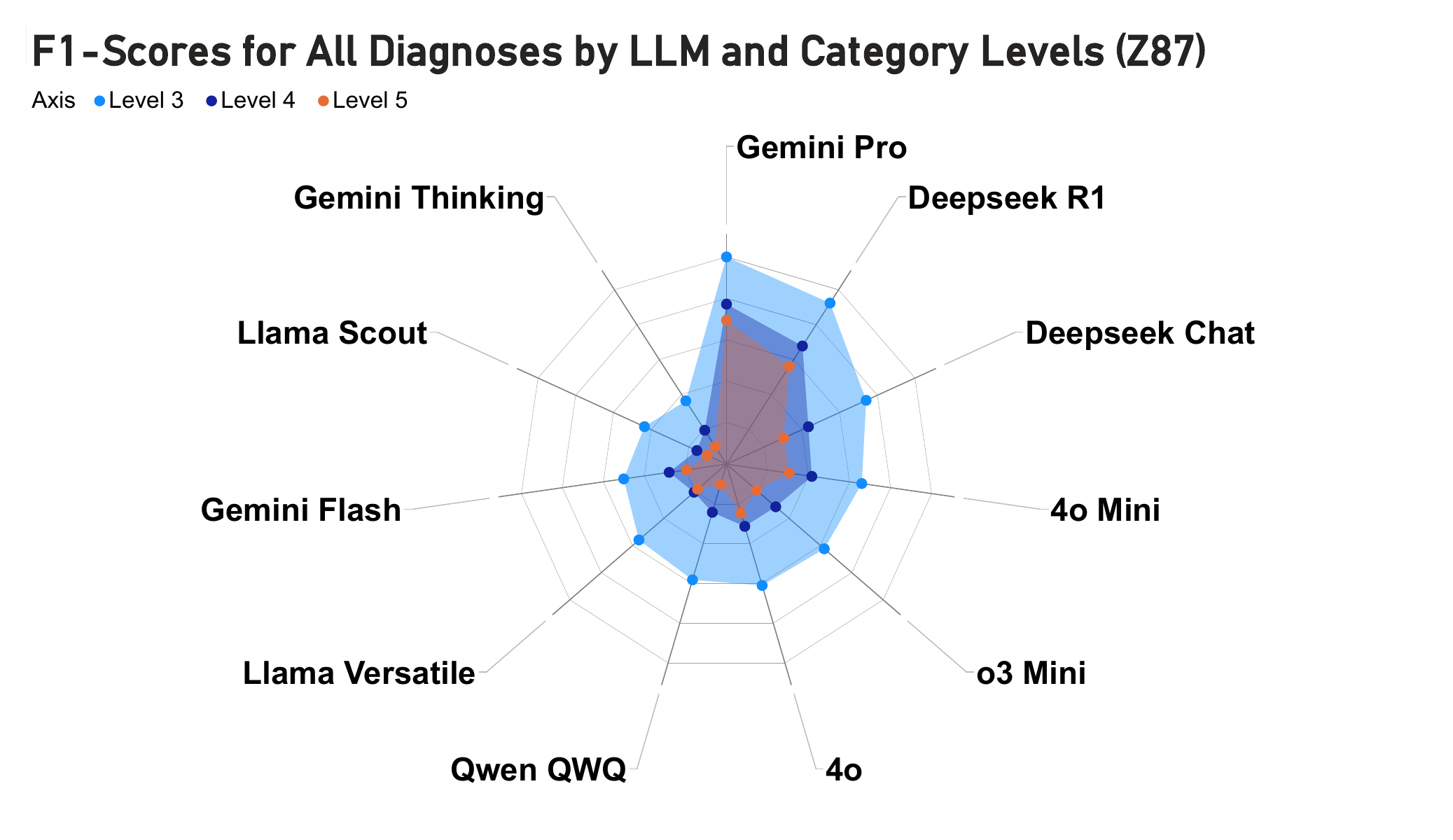}
\noindent
\caption{F1 scores for ICD-10 code Z87 in all diagnoses,\\by LLM, across ICD-10 levels 3 to 5}
\label{MiAllZ87}
\end{subfigure}%
\caption{F1 scores for ICD-10 code Z87, by LLM, across ICD-10 levels 3 to 5}

\end{figure}

\end{appendices}

\end{document}